\RequirePackage{amsthm}
\documentclass[sn-mathphys-ay]{sn-jnl}

\usepackage{graphicx}%
\usepackage{multirow}%
\usepackage{amsmath,amssymb,amsfonts}%
\usepackage{amsthm}%
\usepackage{mathtools}
\usepackage{mathrsfs}%
\usepackage[title]{appendix}%
\usepackage{xcolor}%
\usepackage{textcomp}%
\usepackage{manyfoot}%
\usepackage{booktabs}%
\usepackage{algorithm}%
\usepackage{algorithmicx}%
\usepackage{algpseudocode}%
\usepackage{listings}%
\usepackage[capitalize,noabbrev]{cleveref}
\usepackage[tight-spacing=true]{siunitx}
\usepackage{bm}
\theoremstyle{thmstyleone}%
\theoremstyle{thmstyletwo}%

\theoremstyle{thmstylethree}%

\newtheorem{hyp}{Assumption}

\newcommand{\x}{\mathbf{x}}
\newcommand{\R}{\mathbb{R}}
\newcommand{\N}{\mathbb{N}}

\begin{document}

\title[Hybrid additive modeling with partial dependence for supervised regression and dynamical systems forecasting]{Hybrid additive modeling with partial dependence for supervised regression and dynamical systems forecasting}

\author*[1]{\fnm{Yann} \sur{Claes}}\email{y.claes@uliege.be}
\author[1]{\fnm{V\^an Anh} \sur{Huynh-Thu}}\email{vahuynh@uliege.be}
\author[1]{\fnm{Pierre} \sur{Geurts}}\email{p.geurts@uliege.be}

\affil*[1]{\orgdiv{Montefiore Institute}, \orgname{University of Liège}, \orgaddress{\city{Liège}, \postcode{4000}, \country{Belgium}}}


\abstract{Learning processes by exploiting restricted domain knowledge is an important task across a plethora of scientific areas, with more and more hybrid training methods additively combining data-driven and model-based approaches. Although the obtained models are more accurate than purely data-driven models, the optimization process usually comes with sensitive regularization constraints. Furthermore, while such hybrid methods have been tested in various scientific applications, they have  been mostly tested on dynamical systems, with only limited study about the influence of each model component on global performance and parameter identification. In this work, we introduce a new hybrid training approach based on partial dependence, which removes the need for intricate regularization. Moreover, we assess the performance of hybrid modeling against traditional machine learning methods on standard regression problems. We compare, on both synthetic and real regression problems, several approaches for training such hybrid models. We focus on hybrid methods that additively combine a parametric term with a machine learning term and investigate model-agnostic training procedures. Therefore, experiments are carried out with different types of machine learning models, including tree-based models and artificial neural networks. We also extend our partial dependence optimization process for dynamical systems forecasting and compare it to existing schemes.\footnote{Our Python implementations of the hybrid methods are available at \url{https://github.com/yannclaes/kg-regression}.}}

\keywords{Physics-informed machine learning, Physics-guided machine learning, Supervised regression, Tree-based methods, Neural networks, Partial dependence, Hybrid modeling.}



\maketitle

\section{Introduction}\label{sec:introduction}

For the past decades, machine learning (ML) models have been developed to tackle a variety of real-life problems, complementing/replacing model-based (MB) approaches, which mostly remain approximations that make stringent assumptions about the system under study. Traditional ML approaches are said to be \textit{data-driven}, i.e.\@ their prediction model is solely built from some learning dataset. While their design comes with great expressiveness, they are likely to be subject to over-fitting without enough training examples and to show a lack of robustness on unseen samples, with predictions that can be inconsistent with respect to domain knowledge \citep{daw2022physics,de2019deep}. To overcome this generalization issue, hybrid approaches have been introduced to incorporate \textit{a priori} domain knowledge within statistical models, which can be leveraged in a multitude of ways (see \citealp{von2021informed,Karniadakis21,willard2022integrating} for reviews). A popular hybrid approach considers an additive combination of MB and ML components, and the success of these methods have been shown empirically on a range of synthetic and real-world problems \citep{yin2021augmenting,mehta2021neural,dona2022constrained}.
However, a significant challenge with such methods is to find the right balance between MB and ML components, if one is interested in retrieving a meaningful estimation of the parametric prior. This generally requires the introduction of regularization terms in the loss function, whose weighting coefficients are complex to optimize. Moreover, while these hybrid optimization methods have been mostly applied to dynamical systems, they have not been thoroughly studied in the context of standard regression problems, and the majority of ML models considered in these approaches are neural networks, leaving aside other methods. Our contributions are the following: 
\begin{itemize}
    \item We introduce a new hybrid training method based on partial dependence (PD), which avoids regularization constraints and makes fewer assumptions than other approaches.
    \item We first investigate empirically the performance and benefits of hybrid methods against data-driven methods on \textit{static} regression problems (in contrast to dynamical problems). The static context removes a layer of complexity related to the temporal correlation between observed states, which makes it easier to assess the interaction between the MB and ML components. We compare different approaches for training such hybrid additive models, including our PD-based training. We highlight specific assumptions under which these approaches are expected to work well and relate their differences in terms of prediction and parameter recovery performance. We focus on model-agnostic approaches, where the ML term can be of any type, and we compare tree-based methods against neural networks. Tree-based methods have several advantages over neural networks, which motivate their use on static regression problems: they have much less hyper-parameters to tune, appear robust to the presence of irrelevant features and have been shown to often outperform neural networks on tabular data \citep{grinsztajn2022tree}.
    \item We extend our partial-dependence-based optimization scheme to dynamical systems forecasting and compare it to the other hybrid training approaches. We show that it performs equally well and that it produces parametric physical terms of very good quality compared to other hybrid additive training schemes.
\end{itemize}

{\it\noindent Note: This paper is an extended version of \citep{claes2023}. In addition to improved descriptions of the algorithms and experiments, this paper includes the adaptation and evaluation of the methods for dynamical systems forecasting, while \citep{claes2023} focused exclusively on supervised regression.}

\section{Related work}\label{sec:related_work}
Hybrid additive modeling methods emerged several decades ago, combining first-principles models with different ML models. Already in the 1990's, approaches derived by \citet{kramer1992embedding,johansen1992representing,thompson1994modeling} complemented physics-based models with neural networks, weighting contributions of both components (e.g.\@ through radial basis function networks), to achieve enhanced physical consistency with better generalization properties. More recently, other works applied the same principles to model dynamical systems in various domains, still massively relying on neural networks \citep{yin2021leads,qian2021integrating,takeishi2021physics,wehenkel2022improving}. In a more standard regression setting, \citet{chen2007partially,yu2010partially,zhang2019regression} combined a linear parametric term with a tree-based ML term.

On the challenge of finding the right balance between the MB and ML components, previous works have introduced regularization of the ML term to reduce parameter identification issues in the decomposition \citep{wouwer2004biological,hu2011hybrid}. Further works on this matter introduced constraints in the learning objective to better control contributions of the MB/ML components \citep{yin2021augmenting,dona2022constrained}. Elements of discussion about the well-posedness of this additive decomposition have also been introduced in previous works:  \citet{yin2021augmenting} showed the existence and uniqueness of an optimal MB-ML pair when the contributions of the statistical model are constrained to be minimal, and \citet{dona2022constrained} demonstrated the convergence of an algorithm alternating between the optimization of both parts, without however any guarantee about convergence points. Nevertheless, most of these works only addressed dynamical systems forecasting.

\section{Static supervised regression}

In this section, we focus first on hybrid additive modeling for standard supervised regression. We formally define the problem of hybrid additive modeling in Section \ref{sec:prob_statement} and then describe various methods to address it in Section \ref{sec:methods}, including a new proposal based on partial dependence (Section \ref{sec:PDP}). All methods are compared on several synthetic and real problems in Section \ref{sec:exp_static}.

\subsection{Problem statement}\label{sec:prob_statement}

Let us define a regression problem, where $y \in \R$ and $\x \in \R^d$, with $d \in \N_+$, are drawn from a distribution $p(\x,y)$ such that $y = f(\x) + \varepsilon$
with $f : \R^d \mapsto \R$ the partially known generating function and $\varepsilon \sim \mathcal{N}(0, \sigma^2)$ the noise term. We focus on problems such that $f(\x)$ can be decomposed as:
\begin{hyp}[A1, Additivity]\label{hyp:1} $y = f_k(\x_k) + f_a(\x) + \varepsilon,$ where $\x_k$ is a subset of $K \leq d$ input variables.
\end{hyp}
We assume partial knowledge of the generating function through some known algebraic function $h_k^{\theta_k}(\x_k) \in \mathcal{H}_k$ with tunable parameters $\theta_k$, such that for the optimal parameter values $\theta_k^*$ we have $h_k^{\theta_k^*} = f_k$. The residual term $f_a(\x)$ is unknown and is approximated in this work through an ML component $h_a^{\theta_a} \in \mathcal{H}_a$, with parameters $\theta_a$\footnote{In the following, $h_k^{\theta_k}$ and $h_a^{\theta_a}$ will sometimes be denoted simply as $h_k$ and $h_a$ to lighten the notations.}. The final model $h\in {\cal H}$ is denoted $h(\x) = h_k^{\theta_k}(\x_k) + h_a^{\theta_a}(\x)$, with the function space $\cal H$ defined as ${\cal H}_k + {\cal H}_a$. A\ref{hyp:1} is common when MB methods and ML models are combined \citep{takeishi2021physics,yin2021leads,dona2022constrained,wehenkel2022improving}.

Given a learning sample of $N$ input-output pairs $LS = \{(\x_i, y_i)\}_{i=1}^N$, drawn from $p(\x,y)$, we seek to identify a function $h = h_k^{\theta_k} + h_a^{\theta_a}$, i.e. parameters $\theta_k$ and $\theta_a$, that minimizes the following two distances:
\begin{align}
    d(h, y) &= \mathbb{E}_{(\x,y)\sim p(\x,y)} \{(h(\x) - y)^2\},\label{eq:obj1}\\
    d_k(h^{\theta_k}_k, f_k) &= \mathbb{E}_{\x_k \sim p(\x_k)}\{ (h^{\theta_k}_k(\x_k) - f_k(\x_k))^2\}. \label{eq:obj2}
\end{align} 
The first distance measures the standard generalization error of the global model $h$. The hope is that taking $h_k$ into account will help learning a better global model than fitting directly a pure data-driven model on $y$, especially in the small sample size regime. The second distance $d_k$ measures how well the tuned $h_k$ approximates $f_k$. The main motivation for this second objective is interpretability: one expects that the algebraic form of $h_k$ will be derived from first principles by domain experts, who will be interested in estimating the parameters of this term from data. An alternative to $d_k$ is a loss that would compare the estimated and optimal parameters $\hat{\theta}_k$ and $\theta^*_k$ (e.g., $||\hat{\theta}_k - \theta^*_k||^2$). $d_k$ however has the advantage not to require $\theta_k^*$ to be fully identifiable, i.e.\@ there can exist several sets of parameters $\theta^*_k$ such that $h^{\theta^*_k}_k = f_k$. 

The following approximation of \eqref{eq:obj1} can be used as training objective:
\begin{equation}\label{eq:approx_obj1}
    \hat{d}(h, y; LS) = \frac{1}{N} \sum_{i=1}^N (h(\mathbf{x}_i) - y_i)^2.
\end{equation}
Minimizing the distance in \eqref{eq:obj2} is expected to be challenging and sometimes even ill-posed. Indeed, if $h_a$ is too powerful, it could capture $f$ entirely and leave little room for the estimation of $f_k$. Finding the right balance between $h_k$ and $h_a$ is thus very challenging, if not impossible, using only guidance of the learning sample $LS$. Unlike \eqref{eq:obj1}, \eqref{eq:obj2} cannot be estimated from a sample of input-output pairs and hence cannot be explicitly used to guide model training. There are however several scenarios that can make the problem easier. In the following, we will discuss the optimality of hybrid methods under two additional assumptions:

\begin{hyp}[A2, Disjoint features]\label{hyp:2}
Let $\x_a$ be a subset of features disjoint from $\x_k$ ($\x_k\cap \x_a = \emptyset$). There exists a function $f^r_a(\x_a)$ such that $f_a(\x) = f^r_a(\x_a)$ for all $\x$.
\end{hyp}

\begin{hyp}[A3, Independence]\label{hyp:3}
Features in $\x_k$ are independent from features in $\x_a$ ($\x_k\perp\!\!\!\perp \x_a$).
\end{hyp}

Problems where A\ref{hyp:2} is satisfied are easier as $f_k$ captures all the dependence of $y$ on $\x_k$. Furthermore, in problems where A\ref{hyp:3} is not met, it might be hard to distinguish real contributions from $\x_k$ to $f$ from those due to correlations with features not in $\x_k$. Note that these assumptions do not strictly define our modeling framework, but are rather used in further developments and experiments as tools to derive claims about the optimality and convergence of hybrid training methods on problems that do (or do not) satisfy them.

\subsection{Training methods}\label{sec:methods}

We focus on model-agnostic approaches, i.e.\@ that can be applied with any algebraic function $h_k$ and any type of ML model $h_a$. For both terms, we only assume access to training functions, respectively denoted $\operatorname{fit}^{h_k}$, $\operatorname{fit}^{h_k+\gamma}$, and $\operatorname{fit}^{h_a}$, that can estimate each model parameters, respectively $\theta_k$, $(\theta_k,\gamma)$ and $\theta_a$, so as to minimize the mean squared error (MSE) over $LS$ (see below for the meaning of $\gamma$), where parametric methods rely on gradient descent. Pseudo-codes of methods in Sections \ref{subsec:seq_training_hybrid} and \ref{subsec:alternating_method_general} are given in Appendix \ref{sec:pseudocodes}.

\subsubsection{Sequential training of $h_k$ and $h_a$}
\label{subsec:seq_training_hybrid}

This baseline approach first fits $h_k$ on the observed output $y$, then fits $h_a$ on the resulting residuals, as done by \citet{zhang2019regression} (see \cref{alg:sequential_opt_general_static} in Appendix \ref{sec:pseudocodes}). More precisely, we first train $h_k^{\theta_k}$ on $y$ by introducing a constant term $\gamma \in \mathbb{R}$, such that 
\begin{equation}
\label{eq:hk_only_optimization}
    (\hat{\theta}_k, \hat{\gamma}) = \operatorname{fit}^{h_k + \gamma}(LS).
\end{equation}
Our motivation for introducing the term $\gamma$ will be explained below. Afterwards, we fit $h_a$ on the output residuals: $\hat{\theta}_a = \operatorname{fit}^{h_a}\{(\x_i, y_i - h_k^{\hat{\theta}_k}(\x_i)-\hat{\gamma})\}_{i=1}^N$.

Let $\hat{\mathcal{F}}_k$ be the set of all functions $\hat{f}_k$ mapping $\x_k \in \mathcal{X}_k$ to some value $y \in \R$, i.e. $\hat{\mathcal{F}}_k = \{\hat{f}_k : \mathcal{X}_k \mapsto \R\}$.
Under A\ref{hyp:2} and A\ref{hyp:3}, it can be shown that $\hat{f}_k^* = \arg\min_{\hat{f}_k \in \hat{\mathcal{F}}_k} d(\hat{f}_k, y)$
is such that $\hat{f}_k^*(\x_k) = f_k(\x_k) + C$, for every $\x_k \in \mathcal{X}_k$, with $C = \mathbb{E}_{\x_a}\left\{f^r_a(\x_a)\right\}$ (for more details, see Appendix \ref{app:bayes_model_proof}). Hence, this approach is sound at least asymptotically and justifies the introduction of $\gamma$. Note however that even under A\ref{hyp:2} and A\ref{hyp:3}, we have no guarantee that this approach produces the best estimator for a finite sample size, as $f^r_a(\x_a)+\epsilon$ acts as a pure additive noise term that needs to be averaged out during training. The approaches described in Sections~\ref{subsec:alternating_method_general} and \ref{sec:PDP} try to overcome this issue by fitting $h^{\theta_k}_k$ on corrected outputs that are expected to be closer to $f_k(\x_k)$. Regarding sequential optimization, without A\ref{hyp:2} and A\ref{hyp:3}, the quality of the estimation $h_k^{\hat{\theta}_k}$, according to \eqref{eq:obj2}, is not guaranteed as there are problems satisfying A\ref{hyp:1} such that:
\begin{equation}\label{eq:seqnotop}
\nexists \gamma\in \R: \arg \min_{\hat{f}_k\in \hat{\mathcal{F}}_k} d(\hat{f}_k, y) = f_k + \gamma.
\end{equation}
An example of such problems will be given in Section \ref{subsec:exp_correlated_inputs}.

\subsubsection{Alternate training of $h_k$ and $h_a$}
\label{subsec:alternating_method_general}
A hybrid additive approach was proposed by \citet{dona2022constrained}, tackling the problem of dynamical systems forecasting, which alternates between updating $h_k$ and updating $h_a$, using neural networks for $h_a$. They also introduce specific regularization criteria to try and find the optimal balance between both parts of the model, but we decided to focus on the non-regularized version of the algorithm, as we did not manage to significantly improve the results in the static experiments by adding the regularization process (such process will however have an impact in the dynamical setting). Such alternate training was also proposed by \citet{chen2007partially} and \citet{yu2010partially} with a single decision tree as $h_a$ and a linear $h_k$. We include this approach in our comparison, but also investigate it with random forests \citep{breiman2001random} and tree gradient boosting \citep{friedman2002stochastic}. 

The optimization process defined by \citet{dona2022constrained} is as follows: $\hat{\theta}_k$ is firstly initialized by (fully) fitting $h_k^{{\theta}_k}+\gamma$ on $y$. Then, we alternate between: (1) a single epoch of gradient descent on $h_k^{{\theta}_k}+\gamma$ and (2) either a single epoch for $h_a$ (in the case of neural networks, as performed by \citet{dona2022constrained}) or a complete fit of $h_a$ (in the case of tree-based models). This process is described in \cref{alg:alternating_opt_general_static} (Appendix \ref{sec:pseudocodes}).

While some theoretical results are given by \citet{dona2022constrained}, convergence of the alternate method towards the optimal solution is not guaranteed in general. Despite an initialization favoring $h_k$, it is unclear whether a too expressive $h_a$ will not dominate $h_k$ and finding the right balance between these two terms, e.g.\@ by regularizing further $h_a$, is challenging. Under A\ref{hyp:2} and A\ref{hyp:3} however, the population version\footnote{i.e., assuming an infinite training sample size and consistent estimators.} of the algorithm produces an optimal solution. Indeed, $h_k$ will be initialized as the true $f_k$, as shown previously, making the residuals $y - h_k$ at the first iteration, as well as $h_a$, independent of $\x_k$. $h_k$ will thus remain unchanged (and optimal) at subsequent iterations.

\subsubsection{Partial Dependence-based training of $h_k$ and $h_a$}
\label{sec:PDP}
We propose a novel approach relying on partial dependence (PD) functions \citep{friedman2001greedy} to produce a proxy dataset depending only on $\x_k$ to fit $h_k$. PD measures how a given subset of features impact the prediction of a model, on average. Let $\x_k$ be the subset of interest and $\x_{-k}$ its complement, with $\x_k \cup \x_{-k} = \x$, then the PD of a function $f(\x)$ on $\x_k$ is:
\begin{equation}
\label{eq:pdp_definition}
    PD(f,\x_k) = \mathbb{E}_{\x_{-k}}\left[f(\x_k, \x_{-k})\right] = \int f(\x_k, \x_{-k}) p(\x_{-k}) d\x_{-k}, 
\end{equation}
where $p(\x_{-k})$ is the marginal distribution of $\x_{-k}$. Under A\ref{hyp:1} and A\ref{hyp:2}, the PD of
$f(\x) = f_k(\x_k) + f^r_a(\x_a)$ is \citep{friedman2001greedy}: 
\begin{equation}\label{eq:pdpsound}
PD(f,\x_k) = f_k(\x_k) + C,\mbox{ with } C = E_{\x_a} \{f^r_a(\x_a)\}.
\end{equation}
The idea of our method is to first fit any sufficiently expressive ML model $h_a(\x)$ on $LS$ and to compute its PD w.r.t. $\x_k$ to obtain a first approximation of $f_k(\x_k)$ (up to a constant). Although computing the actual PD using \eqref{eq:pdp_definition} requires in principle access to the input distribution, an approximation can be estimated from $LS$ as follows:
\begin{equation}
    \widehat{PD}(h_a,\x_k; LS) = \frac{1}{N}\sum_{i=1}^{N}h_a(\x_k, \x_{i, -k}),
\end{equation}
where $\x_{i, -k}$ denotes the values of $\x_{-k}$ in the $i$-th sample of $LS$. A new dataset of pairs $(\x_k, \widehat{PD}(h_a,\x_k; LS))$ can then be built to fit $h_k$. In our experiments, we consider only the $\x_k$ values observed in the learning sample but $\widehat{PD}(h_a, \x_k; LS)$ could also be estimated at other points $\x_k$ to artificially increase the size of the proxy dataset. 

In practice, optimizing $\theta_k$ only once on the PD of $h_a$ could leave residual dependence of $\x_k$ on the resulting $y - h_k^{\hat{\theta}_k}(\x_k) - \hat{\gamma}$. We thus repeat the sequence of fitting $h_a$ on the latter residuals, then fitting $h_k$ on the obtained $\widehat{PD}(h_a^{\hat{\theta}_a}, \x_k; LS) + h_k^{\hat{\theta}_k}(\x_k) + \hat{\gamma}$, with $\hat{\theta}_k$, $\hat{\gamma}$ and $\hat{\theta}_a$ the current optimized parameter vectors (see \cref{alg:pdp_opt_general_static}).


\begin{algorithm}
    \caption{Partial Dependence Optimization - Static problems}
    \label{alg:pdp_opt_general_static}
    \begin{algorithmic}
        \State {\bfseries Input:} $LS = \{(\x_i, y_i)\}_{i=1}^N$
        \State $\hat{\theta}_a^{(0)} \leftarrow \operatorname{fit}^{h_a}(LS)$
        \State $(\hat{\theta}_k^{(0)}, \hat{\gamma}^{(0)}) \leftarrow \operatorname{fit}^{h_k+\gamma}(\{(\x_{k,i}, \hspace{3pt} \widehat{PD}(h_a^{\hat{\theta}_a^{(0)}}, \x_{k,i}; LS))\}_{i=1}^N)$
        \For{$n=1$ {\bfseries to} $N_r$}
        \State $\hat{\theta}_a^{(n)} \leftarrow \operatorname{fit}^{h_a}(\{(\x_i, \hspace{3pt} y_i-h_k^{\hat{\theta}_k^{(n-1)}}(\x_{k,i}) - \hat{\gamma}^{(n-1)})\}_{i=1}^N)$
        \State $(\hat{\theta}_k^{(n)}, \hat{\gamma}^{(n)}) \leftarrow \operatorname{fit}^{h_k+\gamma}(\{(\x_{k,i}, \hspace{3pt} h_k^{\hat{\theta}_k^{(n-1)}}(\x_{k,i})+\hat{\gamma}^{(n-1)}+\widehat{PD}(h_a^{\hat{\theta}_a^{(n)}}, \x_{k,i}; LS))\}_{i=1}^N)$
        \EndFor
        \State $\hat{\theta}_a^{(N_r+1)} \leftarrow \operatorname{fit}^{h_a}(\{(\x_i, \hspace{3pt} y_i-h_k^{\hat{\theta}_k^{(N_r)}}(\x_{k,i}) - \hat{\gamma}^{(N_r)})\}_{i=1}^N)$
        \State \Return $h_k^{\hat{\theta}_k^{(N_r)}}+\hat{\gamma}^{(N_r)}+h_a^{\hat{\theta}_a^{(N_r+1)}}$
    \end{algorithmic}
\end{algorithm}

The main advantage of this approach over the alternate one is to avoid domination of $h_a$ over $h_k$. Unlike the two previous approaches, this one is also sound even if A\ref{hyp:3} is not satisfied as it is not a requirement for \eqref{eq:pdpsound} to hold. One drawback is that it requires $h_a$ to capture well the dependence of $f$ on $\x_k$ so that its PD is a good approximation of $f_k$. The hope is that even if it is not the case at the first iteration, fitting $h_k$, that contains the right inductive bias, will make the estimates better and better over the iterations.

\subsection{Experiments}\label{sec:exp_static}
We compare the different methods on several regression datasets, both simulated and real. Performance is measured through estimates of \eqref{eq:obj1} and \eqref{eq:obj2} (the latter only on simulated problems) on a test set $TS$, respectively denoted $\hat{d}(h, y; TS)$ and $\hat{d}_k(h_k^{\theta_k}, f_k; TS)$. In some cases, we also report $\operatorname{rMAE}(\theta_k^*, \theta_k)$, the relative mean absolute error between $\theta_k^*$ and $\theta_k$ (lower is better for all measures). For hybrid approaches, we use as $h_a$ either a multilayer perceptron (MLP), gradient boosting with decision trees (GB) or random forests (RF). We compare these hybrid models to a standard data-driven model that uses only $h_a$. We also compare fitting $h_a$ with and without input filtering, i.e.\@ respectively removing or keeping $\x_k$ from its inputs, to verify convergence claims about $h_k$ in \cref{subsec:alternating_method_general}, under A\ref{hyp:2}. Architectures (e.g.\@ for MLP, the number of layers and neurons) are kept fixed across training methods to allow a fair comparison between them, and are given in Appendix \ref{app:architectures}. We monitor (\ref{eq:approx_obj1}) on a validation set and select the model that reaches the lowest validation loss.

\subsubsection{Independent input features (A\ref{hyp:2} and A\ref{hyp:3} satisfied)} 
\label{subsec:exp_indep_inputs}
\paragraph{Friedman problem}
We consider the following synthetic regression problem:
\begin{equation*}
    y = \theta_{0} \sin(\theta_{1} x_0 x_1) + \theta_{2} (x_2 - \theta_{3})^2 + \theta_{4} x_3 + \theta_{5} x_4 + \sum_{j=5}^{9}0 x_j + \varepsilon,
\end{equation*}
where $x_j \sim \mathcal{U}(0, 1), j=0, \dots 9$, and $\varepsilon \sim \mathcal{N}(0, 1)$ \citep{friedman1983multidimensional}. We generate 10 different datasets using 10 different sets of values for $\theta_0, \dots, \theta_5$, each with 300, 300 and 600 samples for respectively the training, validation and test sets. For the hybrid approaches, we use the first term as prior knowledge, i.e. $f_k(\x_k) = \theta_{0} \sin(\theta_{1} x_0 x_1)$.
\begin{table}
\caption{Results on the \emph{Friedman} problem. We report the mean and standard deviation of $\hat{d}$ and $\hat{d}_k$ over the 10 test sets (TS). ``$f_k \rightarrow h_a$" fits $h_a$ on $y - f_k(\x_k)$. ``Unfiltered" indicates that all the features are used as inputs of $h_a$, while ``Filtered" indicates that the features $\x_k$ are removed from the inputs of $h_a$.}
\label{table:fried1_first_results}
\centering\small
\setlength{\tabcolsep}{8pt}
\begin{tabular}{clcccc}
\toprule
&  & \multicolumn{2}{c}{$\hat{d}(h, y; TS)$} & \multicolumn{2}{c}{$\hat{d}_k(h_k^{\theta_k}, f_k; TS)$}\\
& Method & Unfiltered & Filtered & Unfiltered & Filtered \\ 
\midrule 
& $f_k \rightarrow h_a$ & $1.58 \pm 0.33$ & $1.23 \pm 0.10$ & \multicolumn{2}{c}{-} \\ 
& Sequential & $1.54 \pm 0.31$ & $1.43 \pm 0.13$ & \multicolumn{2}{c}{$0.18 \pm 0.16$} \\ 
MLP & Alternate & $1.43 \pm 0.09$ & $1.32 \pm 0.09$ & $0.10 \pm 0.09$ & $0.02 \pm 0.02$ \\ 
& PD-based & $1.54 \pm 0.12$ & $1.38 \pm 0.09$ & \multicolumn{2}{c}{$0.06 \pm 0.07$} \\ 
& $h_a$ only & \multicolumn{2}{c}{$2.62 \pm 0.75$} & \multicolumn{2}{c}{-} \\ 
\hline
& $f_k \rightarrow h_a$ & $1.73 \pm 0.09$ & $1.75 \pm 0.12$ & \multicolumn{2}{c}{-} \\ 
& Sequential & $1.74 \pm 0.11$ & $1.81 \pm 0.14$ & \multicolumn{2}{c}{$0.18 \pm 0.16$} \\ 
GB & Alternate & $1.79 \pm 0.11$ & $1.78 \pm 0.15$ & $0.91 \pm 1.45$ & $0.06 \pm 0.06$ \\ 
& PD-based & $1.77 \pm 0.13$ & $1.78 \pm 0.12$ & \multicolumn{2}{c}{$0.03 \pm 0.02$} \\ 
& $h_a$ only & \multicolumn{2}{c}{$3.43 \pm 0.94$} & \multicolumn{2}{c}{-} \\ 
\hline
& $f_k \rightarrow h_a$ & $2.03 \pm 0.18$ & $1.96 \pm 0.17$ & \multicolumn{2}{c}{-} \\ 
& Sequential & $2.11 \pm 0.23$ & $2.05 \pm 0.24$ & \multicolumn{2}{c}{$0.18 \pm 0.16$} \\ 
RF & Alternate & $2.03 \pm 0.19$ & $1.98 \pm 0.17$ & $0.04 \pm 0.03$ & $0.04 \pm 0.04$ \\ 
& PD-based & $2.16 \pm 0.27$ & $2.09 \pm 0.26$ & \multicolumn{2}{c}{$0.16 \pm 0.15$} \\
& $h_a$ only & \multicolumn{2}{c}{$5.58 \pm 1.91$} & \multicolumn{2}{c}{-} \\ 
\bottomrule
\end{tabular}
\end{table}

We see in \cref{table:fried1_first_results} that all hybrid training schemes outperform their data-driven counterpart. They come very close to the ideal $f_k \rightarrow h_a$ method, and sometimes even slightly better, probably due to chance. Sequential fitting of $h_k$ and $h_a$ performs as well as the alternate or PD-based approaches, as A\ref{hyp:2} and A\ref{hyp:3} are satisfied for this problem (see \cref{subsec:seq_training_hybrid}). Filtering generally improves the performance of hybrid schemes as A\ref{hyp:2} is satisfied. PD-based optimization yields good approximations of $f_k$ (as shown by a low $\hat{d}_k$). The alternate approach follows closely whereas the sequential one ends up last, which can be expected as fitting $h_k$ only on $y$ induces a higher noise level centered around $\mathbb{E}_{\x_a}\left\{f_a(\x_a)\right\}$, while the other approaches benefit from reduced perturbations through $h_a$ estimation, as explained in \cref{subsec:seq_training_hybrid}. Filtering vastly decreases $\hat{d}_k$ for alternate approaches, supporting claims introduced in \cref{subsec:alternating_method_general}, while this measure remains unimpaired for sequential and PD-based training by construction.

\subsubsection{Correlated input features (A\ref{hyp:3} not satisfied)}
\label{subsec:exp_correlated_inputs}
\paragraph{Correlated Linear problem.}
Let $y = \beta_0 x_0 + \beta_1 x_1 + \varepsilon$, with $\beta_0 = -0.5, \beta_1 = 1, \x \sim \mathcal{N}(\mathbf{0}, \Sigma),$ and $\varepsilon \sim \mathcal{N}(0, 0.5^2)$. We generate 50, 50 and 600 samples respectively for the training, validation and test sets. We use as known term $f_k(\x_k) = \beta_0 x_0$. Regressing $y$ on $x_0$ yields the least-squares solution \citep{greene2018econometric}:
\begin{equation}
\label{eq:biased_beta}
    \mathbb{E}\left[\hat{\beta}_0\right] = \beta_0 + \frac{\operatorname{cov}(x_0, x_1)}{\operatorname{var}(x_0)}\beta_1.
\end{equation}
We set $\operatorname{cov}(x_0, x_1) = 2.25$ and $\operatorname{var}(x_0) = 2$ so that \eqref{eq:biased_beta} reverses the sign of $\beta_0$ and (\ref{eq:seqnotop}) is satisfied. The sequential approach should hence yield parameter estimates of $\beta_0$ close to \eqref{eq:biased_beta} while we expect the others to correct for this bias.
\begin{table}
\caption{Results for the \emph{Correlated Linear} problem. We report $\hat{d}$ and $\operatorname{rMAE}(\beta_0^*, \hat{\beta}_0)$, over 10 different datasets.}
\label{table:linear_data_results}
\centering\small
\setlength{\tabcolsep}{8pt}
\begin{tabular}{clcccc}
\toprule
& & \multicolumn{2}{c}{$\hat{d}(h, y; TS)$} & \multicolumn{2}{c}{$\operatorname{rMAE}(\beta_0^*, \hat{\beta}_0)$}\\
& Method & Unfiltered & Filtered & Unfiltered & Filtered\\
\hline
& Sequential & $0.30 \pm 0.03$ & $0.74 \pm 0.09$ & \multicolumn{2}{c}{$224.14 \pm 13.48$} \\ 
MLP & Alternate & $0.30 \pm 0.02$ & $0.31 \pm 0.04$ & $186.65 \pm 21.31$ & $15.53 \pm 13.57$ \\
& PD-based & $0.30 \pm 0.03$ & $0.29 \pm 0.02$ & \multicolumn{2}{c}{$26.47 \pm 17.32$} \\ 
\hline
& Sequential & $0.59 \pm 0.06$ & $1.38 \pm 0.11$ & \multicolumn{2}{c}{$224.14 \pm 13.48$} \\ 
GB & Alternate & $0.57 \pm 0.06$ & $0.60 \pm 0.09$ & $148.75 \pm 67.35$ & $24.58 \pm 12.20$ \\ 
& PD-based & $0.56 \pm 0.05$ & $0.64 \pm 0.13$ & \multicolumn{2}{c}{$36.05 \pm 17.50$} \\ 
\hline
&Sequential & $0.53 \pm 0.05$ & $0.90 \pm 0.07$ & \multicolumn{2}{c}{$224.14 \pm 13.48$} \\ 
RF & Alternate & $0.43 \pm 0.04$ & $0.42 \pm 0.04$ & $111.04 \pm 52.78$ & $45.38 \pm 22.39$ \\ 
&PD-based & $0.41 \pm 0.03$ & $0.43 \pm 0.04$ & \multicolumn{2}{c}{$57.47 \pm 15.55$} \\ 
\bottomrule
\end{tabular}
\end{table}

From \cref{table:linear_data_results}, we observe that, contrary to PD-based training, the sequential and alternate methods return very bad estimations of $\beta_0$, as A\ref{hyp:3} is no longer satisfied.
Filtering corrects the bias for the alternate approach but degrades the MSE performance for sequential training as it removes the ability to compensate for the $h_k$ misfit.

\paragraph{Correlated Friedman problem.}
The structure is identical to the one in  \cref{subsec:exp_indep_inputs} but with correlated inputs drawn from a multivariate normal distribution where $\mu_i = 0.5$ and $\operatorname{var}(x_i) =  0.75, \forall i$, and $\operatorname{cov}(x_i, x_j) = \pm 0.3, \forall i \neq j$ (the covariance sign being chosen randomly). Sizes of the training, validation and test sets are identical to those of \cref{subsec:exp_indep_inputs}. Inputs are then scaled to be roughly in $[-1, 1]$. Here again, we use $f_k(\x_k) = \theta_0 \sin(\theta_1 x_0 x_1)$.
\begin{table}
\caption{Results for the \emph{Correlated Friedman} problem.}
\label{table:corr_fried1_first_results}
\centering\small
\setlength{\tabcolsep}{8pt}
\begin{tabular}{clcccc}
\toprule
&  & \multicolumn{2}{c}{$\hat{d}(h, y; TS)$} & \multicolumn{2}{c}{$\hat{d}_k(h_k^{\theta_k}, f_k; TS)$}\\
& Method & Unfiltered & Filtered & Unfiltered & Filtered \\ 
\midrule 
& $f_k \rightarrow h_a$ & $1.64 \pm 0.23$ & $1.51 \pm 0.17$ & \multicolumn{2}{c}{-} \\ 
& Sequential & $2.07 \pm 0.40$ & $2.68 \pm 1.38$ & \multicolumn{2}{c}{$1.35 \pm 1.42$} \\ 
MLP & Alternate & $1.95 \pm 0.33$ & $1.62 \pm 0.24$ & $0.49 \pm 0.44$ & $0.14 \pm 0.19$ \\ 
& PD-based & $2.24 \pm 0.31$ & $1.78 \pm 0.30$ & \multicolumn{2}{c}{$0.17 \pm 0.23$} \\ 
& $h_a$ only & \multicolumn{2}{c}{$2.77 \pm 0.73$} & \multicolumn{2}{c}{-} \\ 
\hline
& $f_k \rightarrow h_a$ & $2.58 \pm 0.45$ & $2.53 \pm 0.44$ & \multicolumn{2}{c}{-} \\ 
& Sequential & $2.90 \pm 0.39$ & $3.91 \pm 1.49$ & \multicolumn{2}{c}{$1.35 \pm 1.42$} \\ 
GB & Alternating & $2.67 \pm 0.38$ & $2.62 \pm 0.43$ & $0.51 \pm 0.53$ & $0.22 \pm 0.25$ \\ 
& PD-based & $2.54 \pm 0.35$ & $2.47 \pm 0.36$ & \multicolumn{2}{c}{$0.03 \pm 0.02$} \\ 
& $h_a$ only & \multicolumn{2}{c}{$4.49 \pm 0.66$} & \multicolumn{2}{c}{-} \\ 
\hline
& $f_k \rightarrow h_a$ & $3.02 \pm 0.45$ & $2.93 \pm 0.45$ & \multicolumn{2}{c}{-} \\ 
& Sequential & $3.78 \pm 0.78$ & $4.04 \pm 1.30$ & \multicolumn{2}{c}{$1.35 \pm 1.42$} \\ 
RF & Alternating & $3.06 \pm 0.39$ & $2.99 \pm 0.38$ & $0.14 \pm 0.16$ & $0.15 \pm 0.18$ \\ 
& PD-based & $3.24 \pm 0.38$ & $3.16 \pm 0.37$ & \multicolumn{2}{c}{$0.27 \pm 0.20$} \\ 
& $h_a$ only & \multicolumn{2}{c}{$6.70 \pm 1.47$} & \multicolumn{2}{c}{-} \\ 
\bottomrule
\end{tabular}
\end{table}

As in \cref{subsec:exp_indep_inputs}, \cref{table:corr_fried1_first_results} shows that hybrid models outperform their data-driven equivalents. PD-based methods usually yield more robust $h_k$ estimations in the general unfiltered case, but struggle to line up with the alternate scheme in terms of predictive performance, except for GB-related models. For RF, this can be explained by a worse $h_k$ estimation while for MLP we assume that it is due to $h_a$ overfitting: in the alternate approach, it is optimized one epoch at a time, interleaved with one step on $h_k$, whereas that of PD-based methods is fully optimized (with identical complexities). Sequential and alternate approaches yield poor $h_k$ estimations without filtering since A\ref{hyp:3} is not met, but the latter mitigates this w.r.t. the former, as was already observed in \cref{subsec:exp_indep_inputs}. Input filtering degrades predictive performance for the sequential methods as they cannot counterbalance a poor $h_k$.

\subsubsection{Overlapping additive structure (A\ref{hyp:2} and A\ref{hyp:3} not satisfied)} 
Let $y = \beta x_0^2 + \sin(\gamma x_0) + \delta x_1 + \varepsilon$ with $\varepsilon \sim \mathcal{N}(0, 0.5^2), \beta = 0.2, \gamma = 1.5, \delta = 1$ and $\x$ sampled as in the \emph{Correlated Linear} problem. We generate 50, 50 and 600 samples respectively for the training, validation and test sets. We define $f_k(\x_k) = \beta x_0^2$ and $f_a(\x) = \sin(\gamma x_0) + \delta x_1 + \varepsilon$. Hence, A\ref{hyp:2} and A\ref{hyp:3} do not hold. Even with $\hat{\beta} = \beta^*$,
$h_a$ still needs to compensate for $\sin(\gamma x_0)$. Therefore, filtering is expected to degrade performance for all hybrid approaches as $h_a(x_1)$ will never compensate for this gap, which we observe in \cref{table:overlap_results_mse_fp_mse},  for both MLP and GB. Interestingly, sequential optimization seems to be the most impacted by filtering. It thus seems that having an interaction between the training of $h_k$ and the training of $h_a$ can still help to reduce the generalization error, even though $f_k(x_0)$ (and hence $h_k(x_0)$) is misspecified by design. Results for RF are not shown for the sake of space, but are similar to GB.

\begin{table}
\caption{Results for the \emph{Overlapping} problem.}
\label{table:overlap_results_mse_fp_mse}
\centering\small
\setlength{\tabcolsep}{8pt}
\begin{tabular}{lcccc}
\toprule
& \multicolumn{2}{c}{$\hat{d}(h, y; TS)$} & \multicolumn{2}{c}{$\hat{d}(h, y; TS)$} \\
 & Unfiltered & Filtered & Unfiltered & Filtered \\ 
Method & \multicolumn{2}{c}{MLP} & \multicolumn{2}{c}{GB}\\
\midrule 
$f_k \rightarrow h_a$ & $0.35 \pm 0.02$ & $0.54 \pm 0.04$ & $0.51 \pm 0.04$ & $1.00 \pm 0.12$  \\ 
Sequential & $0.35 \pm 0.01$ & $0.59 \pm 0.05$ & $0.55 \pm 0.07$ & $1.07 \pm 0.11$ \\ 
Alternate & $0.35 \pm 0.02$ & $0.56 \pm 0.05$ & $0.54 \pm 0.09$ & $1.01 \pm 0.11$ \\ 
PD-based & $0.34 \pm 0.02$ & $0.56 \pm 0.05$ & $0.53 \pm 0.05$ & $0.99 \pm 0.12$ \\ 
$h_a$ only & \multicolumn{2}{c}{$0.37 \pm 0.02$} & \multicolumn{2}{c}{$0.55 \pm 0.07$} \\ 
\bottomrule
\end{tabular}
\end{table}

\subsubsection{Impact of training sample size}
\label{sec:training_size}
In this experiment, we want to estimate the performance gap between hybrid methods and data-driven methods, for different sizes of the training sample. We measure the evolution of \eqref{eq:obj1} and \eqref{eq:obj2} for the different training algorithms, for increasing sample sizes of the \emph{Correlated Friedman} problem. In particular, for a fixed training size, we generate 10 datasets using 10 different sets of values for $\theta_0, \dots, \theta_5$. The same 10 validation and test samples are fixed across all training sample sizes, containing 300 and 600 samples each.

\begin{figure}
    \centering
    \includegraphics[width=\linewidth]{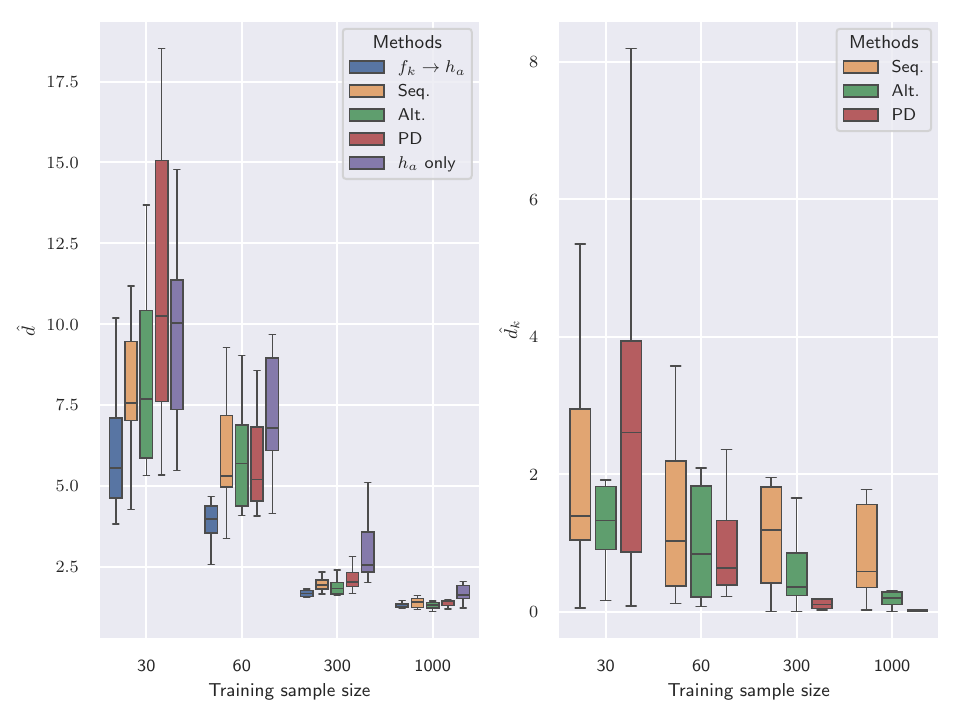}
    \vspace{-2ex}
    \caption{Evolution of $\hat{d}$ (left) and $\hat{d}_k$ (right) w.r.t. the training sample size, on the \emph{Correlated Friedman} problem. The boxplot summarizes the results over 10 test sets, using MLP as $h_a$.}
    \label{fig:impact_ts_mlp}
\end{figure}

We observe in \cref{fig:impact_ts_mlp} that the theoretically ideal $f_k \rightarrow h_a$ method generally performs better than any other training scheme, and significantly better for low-size regimes. Furthermore, hybrid models almost always outperform data-driven methods, except for PD-based training which struggles with very small sample sizes (30 training samples). This could be explained by the fact that, in this algorithm, the training of $h_k$ highly depends on the quality of fit of $h_a$, which serves as surrogate model for the true function $f$. Nevertheless, with slightly more training samples, we notice that it performs as good as the other training schemes. As could be expected, all training methods benefit from an increasing training sample size, and the gap between data-driven and hybrid models becomes smaller.

As observed for the generalization error, we also observe that the PD-based optimization yields worse estimates of $h_k$ compared to the other hybrid approaches for very small sample sizes.  However, it becomes competitive with the other methods at training samples of size 60, and eventually outperforms them at larger sample sizes.

Similar experiments have been conducted using RF and boosting with trees, and are reported in Appendix \ref{app:training_size_additional}.

\subsubsection{Real regression problems}
We now apply all methods on two real-world static datasets. As algebraic term $h_k$, we chose to use a linear prior on $x_k$, where $x_k$ is the feature with the highest importance score in a RF model trained on the full dataset (assumed to be inaccessible at training time). As there is no guarantee that any of our assumptions are met (and in particular A\ref{hyp:1}), we do not measure the distance $d_k$ in \eqref{eq:obj2} as it is deemed irrelevant.

We consider two settings for each dataset, inspired by \citet{zhang2019regression}. In the first setting (INT), the training and test sets are sampled from the same distribution $p(\x, y)$ whereas the second one (EXT) evaluates extrapolation performance for samples with unseen target values. If the linear prior is a reasonable assumption or, at the very least, the target increases or decreases monotonically with $x_k$, then we can expect hybrid methods to yield better results in the latter setting. 
In the INT setting for each dataset, we randomly select 100 samples for the learning set, 100 samples for the validation set and keep the rest as test set. For the EXT setting, we select the samples (one fourth of the dataset) with the lowest output values as test set. From the remaining samples, we randomly select 100 samples for the learning set and 100 samples for the validation set. For both INT and EXT settings, performance metrics are averaged over 10 different splits. We standardize both input and output variables.

The features for both datasets are described in \cref{table:variables_description}. The {\it Combined Cycle Power Plant} (CCPP) dataset \citep{misc_combined_cycle_power_plant_294} collects 9,568 measurements of net hourly electrical energy output for a combined cycle power plant, along with four hourly average input variables. The {\it Concrete Compressive Strength  (CCS) dataset \citep{misc_concrete_compressive_strength_165}} is composed of 1,030 samples relating amounts of concrete components with the resulting compressive strength. As done in the work of \citet{zhang2019regression}, we introduce a new feature corresponding to the cement-to-water ratio.

\begin{table}
    \caption{Variables used for the real-world datasets. For each dataset, the variable indicated in bold type is the one used in the linear prior ($x_k$).}
    \label{table:variables_description}
    \centering
    \begin{tabular}{lll}
    \toprule
    Dataset & Name & Description\\
    \midrule
    CCPP & {\bf T} & {\bf Ambient temperature [°C]}\\
    & AP & Ambient pressure [mbar]\\
    &RH & Relative humidity [-]\\
    & V & Exhaust vacuum [cmHg]\\
    \midrule
    CCS & Cement & Amount of cement in the mixture $\operatorname{[kg/m^3]}$\\
    & Blast Furnace Slag & Amount of blast furnace slag in the mixture $\operatorname{[kg/m^3]}$ \\
    & Fly Ash & Amount of fly ash in the mixture $\operatorname{[kg/m^3]}$ \\
    & Water & Amount of water in the mixture $\operatorname{[kg/m^3]}$ \\
    & Superplasticizer & Amount of superplasticizer in the mixture $\operatorname{[kg/m^3]}$ \\
    & Coarse Aggregate & Amount of coarse aggregate in the mixture $\operatorname{[kg/m^3]}$ \\
    & Fine Aggregate  & Amount of fine aggregate in the mixture $\operatorname{[kg/m^3]}$ \\
    & Age & Day (1-365) \\
    & {\bf Cement/Water} & {\bf Cement to water ratio [-]}\\
    \bottomrule
    \end{tabular}
\end{table}
\begin{table}
\caption{Results for the real-world datasets. We report the mean and standard deviation of $\hat{d}$ over the test sets.}
\label{table:real_world_mse}
\centering\small
\setlength{\tabcolsep}{8pt}
\begin{tabular}{clcccc}
\toprule
& & \multicolumn{2}{c}{CCPP} & \multicolumn{2}{c}{CCS}\\
& Method & INT & EXT & INT & EXT \\ 
\midrule 
& Sequential & $0.07 \pm 0.01$ & $0.23 \pm 0.13$ & $0.25 \pm 0.04$ & $0.74 \pm 0.24$ \\ 
MLP & Alternating & $0.07 \pm 0.01$ & $0.24 \pm 0.10$ & $0.24 \pm 0.03$ & $0.89 \pm 0.36$ \\ 
& PD-based & $0.07 \pm 0.01$ & $0.07 \pm 0.01$ & $0.25 \pm 0.03$ & $0.38 \pm 0.04$ \\ 
& $h_a$ only & $0.07 \pm 0.01$ & $0.36 \pm 0.21$ & $0.25 \pm 0.05$ & $0.83 \pm 0.30$ \\ 
\midrule 
& Sequential & $0.08 \pm 0.01$ & $0.35 \pm 0.07$ & $0.21 \pm 0.02$ & $0.91 \pm 0.24$ \\ 
GB & Alternating & $0.08 \pm 0.01$ & $0.41 \pm 0.17$ & $0.20 \pm 0.03$ & $0.85 \pm 0.28$ \\ 
& PD-based & $0.08 \pm 0.02$ & $0.10 \pm 0.01$ & $0.22 \pm 0.02$ & $0.31 \pm 0.06$ \\ 
& $h_a$ only & $0.10 \pm 0.01$ & $0.95 \pm 0.18$ & $0.25 \pm 0.02$ & $1.54 \pm 0.28$ \\ 
\midrule
& Sequential & $0.07 \pm 0.01$ & $0.28 \pm 0.07$ & $0.20 \pm 0.02$ & $1.06 \pm 0.31$ \\ 
RF & Alternating & $0.07 \pm 0.01$ & $0.32 \pm 0.12$ & $0.20 \pm 0.02$ & $1.00 \pm 0.32$ \\ 
& PD-based & $0.07 \pm 0.01$ & $0.08 \pm 0.01$ & $0.20 \pm 0.02$ & $0.30 \pm 0.08$ \\ 
& $h_a$ only & $0.08 \pm 0.01$ & $0.99 \pm 0.19$ & $0.25 \pm 0.02$ & $1.89 \pm 0.39$ \\
\bottomrule
\end{tabular}
\end{table}

From \cref{table:real_world_mse}, it seems that introducing a linear prior does not yield any benefit in the interpolation setting, as all models perform equally well, which suggests that either the prior is not adequate or that A\ref{hyp:1} is not satisfied. In the extrapolation scenario, we can however observe that the linear prior allows to mitigate the impact of moving out of the distribution compared to data-driven models.
Indeed, compared to the INT setting, the performance of all purely data-driven methods degrades in the EXT scenario, especially for GB and RF as their output predictions are bounded by the minimum target value observed in the training set. PD-based hybrid methods consistently outperform other hybrid approaches and are only slightly impacted, while sequential and alternating methods attain similar results.

\section{Dynamical systems forecasting}
In this section, we focus on the problem of dynamical systems forecasting. We start by defining more formally the problem in \cref{sec:dyn_pb_statement}. In \cref{subsec:dyn_methods}, we explain how PD-based training can be extended to the dynamical system setting and we present the hybrid optimization processes that it will eventually be compared to on several physical systems in \cref{sec:dyn_experiments}.

\subsection{Problem statement}\label{sec:dyn_pb_statement}
In this section, we target the problem of dynamical systems forecasting, i.e. predicting the evolution of systems whose dynamics are driven by:
\begin{equation}
    \dot{\x}_t \coloneq \frac{d\x_t}{dt} = f(\x_t),
\end{equation}
where the state $\x_t$ can be vector-valued, i.e. $\x_t \in \mathbb{R}^d$, or a $d$-dimensional vector field defined over some spatial domain, and $f(\x_t)$ has the same dimension as $\x_t$. We assume access to a finite set $LS$ of $N$ observed state trajectories over a given time interval $[0, T]$, with $T$ the time horizon, i.e. $LS = \{\x_{0}^{(i)}, \dots, \x_{T}^{(i)}\}_{i=1}^N$. For clarity of notation, we use a single $T$ although the time horizon can vary according to the considered system and between the $N$ trajectories. Specifically, we consider systems whose dynamics $f$ can be additively decomposed so as to satisfy A\ref{hyp:1}, i.e.:
\begin{equation}
    f(\x_t) = f_k(\x_{k, t}) + f_a(\x_t).
\end{equation}
As in \cref{sec:prob_statement}, we assume partial knowledge through some parametric term $h_k^{\theta_k} \in \mathcal{H}_k$ with trainable parameters $\theta_k$, such that for the optimal $\theta_k^*$ we have $h_k^{\theta_k^*} = f_k$. The residual function $f_a$ is modeled by some ML component $h_a^{\theta_a}$, such that the final model is given by $h = h_k + h_a$. Similarly to the static setting, we introduce a constant term $\gamma$ to the physical term $h_k^{\theta_k}$, such that we fit $h_k^{\theta_k} + \gamma$. 

We seek to minimize the following two distances:
\begin{align}
    d(h, f) &= \mathbb{E}_{\x \sim p(\x)} ||h(\x) - f(\x)||^2,\label{eq:mse_dynamic}\\
    d_k(h^{\theta_k}_k, f_k) &= \mathbb{E}_{\x \sim p(\x)}\{ (h^{\theta_k}_k(\x_k) - f_k(\x_k))^2\}, \label{eq:mse_hk_dynamic}
\end{align} 
where $p(\x)$ is the distribution over states. The first one measures the distance between the true dynamics $f$ and its approximation $h$. The second distance $d_k$ measures how well the tuned $h_k$ approximates $f_k$. In practice, minimizing (\ref{eq:mse_dynamic}) translates into minimizing the distance between observed trajectories and trajectories obtained by integration of $h$:
\begin{equation}\label{eq:approx_mse_dynamic}
    \hat{d}(h, f; LS) = \frac{1}{N}\frac{1}{T}\sum_{i=1}^{N}\sum_{t=1}^{T}(\hat{\x}_t^{(i)} - \x_t^{(i)})^2,
\end{equation}
where $\hat{\x}_t^{(i)}$ is the solution of the integration of $h$ with initial condition $x_0^{(i)}$ up to time $t$. In our experiments, we measure (\ref{eq:mse_hk_dynamic}) on some test set TS as:
\begin{equation}\label{eq:approx_mse_hk_dynamic}
    \hat{d}(h_k, f_k; TS) = \frac{1}{N}\frac{1}{T}\sum_{i=1}^{N}\sum_{t=1}^{T}(h_k(\x_{k,t}^{(i)}) - f_k(\x_{k,t}^{(i)}))^2.
\end{equation}

\subsection{Training methods}\label{subsec:dyn_methods}
Extending the algorithms of \cref{sec:methods} to dynamical problems requires the integration of $h$ to obtain the solutions $\hat{\x}_t^{(i)}$, which we do through the Neural ODE approach \citep{chen2018neuralode}. Contrary to the static experiments, we solely relied on ML models that allow backpropagation through time (allowing for easy integration with Neural ODE), thereby excluding tree-based models. For that reason, we can now introduce another hybrid training scheme in the comparisons, namely joint optimization of $h_k$ and $h_a$, that was proposed by \citet{yin2021augmenting}. Note that this optimization algorithm could not be used in the static experiments due to the fact that tree-based methods cannot be trained jointly with the parametric component $h_k$.

Given the model approximation error observed on $h_k$ for the sequential scheme in previous results (see \cref{table:linear_data_results} and \cref{table:corr_fried1_first_results}), and assuming that most dynamical systems exhibit correlation between $f_k$ and $f_a$ through time, we choose to omit the sequential optimization scheme and find filtering irrelevant in such a configuration. Hence, we compare our PD-based hybrid scheme with the joint and alternate optimization of $h_k$ and $h_a$, as defined by \citet{yin2021augmenting} and \citet{dona2022constrained} respectively. As for the static case, we compare all schemes to the theoretically best setting of fitting $h_a$ on $f - f_k$, i.e., assuming that the algebraic term if perfectly known. 

\subsubsection{Joint optimization}
In their work, \citet{yin2021augmenting} optimize $h_k^{\theta_k}$ and $h_a^{\theta_a}$ jointly to minimize \cref{eq:mse_dynamic}. They introduce a regularizing criterion on $h_a$ such that its contribution to the total model $h_k + h_a$ is kept minimal, ensuring existence and uniqueness of the minimizing pair $(h_k, h_a)$. We describe the non-regularized joint optimization process in \cref{alg:joint_opt_general} (Appendix \ref{sec:pseudocodes}). More details about the regularizer can be found in \citep{yin2021augmenting}.

\subsubsection{Alternate optimization}
As described in \cref{subsec:alternating_method_general}, $(\hat{\theta}_k, \hat{\gamma})$ is initialized as the minimizer of $d(h_k^{\theta_k}, f)$, then this optimization scheme extends the previous one by alternating training steps between $h_k$ and $h_a$. \citet{dona2022constrained} introduce an additional constraint on $h_k$ to keep it close to its initialization in early training. The non-regularized training scheme is described in \cref{alg:alternating_opt_general_dynamic} (Appendix \ref{sec:pseudocodes}). Details about the regularizer can be found in \citep{dona2022constrained}.

\subsubsection{Partial-dependence-based optimization}
Our optimization scheme can be easily extended to the dynamical setting (see \cref{alg:pdp_opt_general_dynamic}). We first fit any ML model $h_a^{\theta_a}$ on observed trajectories with Neural ODE, which we denote $\operatorname{fit}_{\operatorname{NODE}}^{h_a}$. At this stage, $h_a^{\hat{\theta}_a}$ models the complete system dynamics $f$, hence it can be evaluated for multiple $\x_{k,t}$ values to obtain an initial approximation of $f_k(\x_{k,t})$ through $\widehat{PD}(h_a,\x_{k,t}; LS)$, on which we can fit $h_k^{\theta_k} + \gamma$ as for static regression problems, yielding an initial approximation $h_k^{\hat{\theta}_k} + \hat{\gamma}$. Due to potential residual dependencies, we then alternate between several epochs of fitting $h_a^{\theta_a}$ with Neural ODE to minimize $d(h_a^{\theta_a} + h_k^{\hat{\theta}_k} + \hat{\gamma}, f)$ (keeping $(\hat{\theta}_k, \hat{\gamma})$ fixed) and several epochs of fitting $h_k^{\theta_k} + \gamma$ on $\widehat{PD}(h_a^{\hat{\theta}_a},\x_{k,t}; LS) + h_k^{\hat{\theta}_k} + \hat{\gamma}$, with $\hat{\theta}_k$, $\hat{\gamma}$ and $\hat{\theta}_a$ the current parameter vectors.

\begin{algorithm}
    \caption{Partial Dependence Optimization - Dynamical problems}
    \label{alg:pdp_opt_general_dynamic}
    \begin{algorithmic}
        \State {\bfseries Input:} $LS = \{(\x_0^{(i)}, \dots, \x_T^{(i)})\}_{i=1}^N$
        \State $\hat{\theta}_a^{(0)} \leftarrow \operatorname{fit}_{\operatorname{NODE}}^{h_a}(LS; h_k=0)$
        \State $(\hat{\theta}_k^{(0)}, \hat{\gamma}^{(0)}) \leftarrow \operatorname{fit}^{h_k+\gamma}(\{(\x_{k,t}^{(i)}, \hspace{3pt} \widehat{PD}(h_a^{\hat{\theta}_a^{(0)}}, \x_{k,t}^{(i)}; LS))\}_{i,t=1}^{N, T})$
        \For{$n=1$ {\bfseries to} $N_{r}$}
        \State $\hat{\theta}_a^{(n)} \leftarrow \operatorname{fit}_{\operatorname{NODE}}^{h_a}(LS; h_k=h_k^{\hat{\theta}_k^{(n-1)}}+\hat{\gamma}^{(n-1)})$
        \State $(\hat{\theta}_k^{(n)}, \hat{\gamma}^{(n)}) \leftarrow \operatorname{fit}^{h_k+\gamma}(\{(\x_{k,t}^{(i)}, \hspace{3pt} h_k^{\hat{\theta}_k^{(n-1)}}(\x_{k,t}^{(i)})+\hat{\gamma}^{(n-1)}+\widehat{PD}(h_a^{\hat{\theta}_a^{(n)}}, \x_{k,t}^{(i)}; LS))\}_{i,t=1}^{N,T})$
        \EndFor
        \State $\hat{\theta}_a^{(N_r+1)} \leftarrow \operatorname{fit}_{\operatorname{NODE}}^{h_a}(LS; h_k=h_k^{\hat{\theta}_k^{(N_r)}}+\hat{\gamma}^{(N_r)})$
        \State \Return $h_k^{\hat{\theta}_k^{(N_r)}}+\hat{\gamma}^{(N_r)}+h_a^{\hat{\theta}_a^{(N_r+1)}}$
    \end{algorithmic}
\end{algorithm}

\subsection{Experiments}\label{sec:dyn_experiments}
We compare the different training methods on three simulated dynamical systems that were used for evaluating related methods \citep{yin2021augmenting, dona2022constrained}. We measure performance through log-estimates of \eqref{eq:approx_mse_dynamic} and estimates of \eqref{eq:approx_mse_hk_dynamic} on a test set $TS$, which we respectively denote $\log$-$\hat{d}(h, f; TS)$ and $\hat{d}_k(h_k^{\theta_k}, f_k; TS)$ (lower is better). For hybrid approaches, we use as $h_a$ either a multilayer perceptron (MLP) or a convolutional neural network (CNN), depending on the system. We compare these hybrid models to a standard data-driven framework that uses only $h_a$. Architectures (e.g.\@ for MLP, the number of layers and neurons) are optimized for each training method (unless clearly specified) on the validation set. Details about the hyperparameter optimization procedure and their optimal values are given in Appendix \ref{app:architectures}. We monitor (\ref{eq:approx_mse_dynamic}) on a validation set and select the model that reaches the lowest validation loss. Datasets and optimization details are provided in Appendix \ref{app:dynamical_details}. For conciseness, we only report results obtained using the optimization scheme as exactly described in each reference work and compare the impact of using the specific regularization criterion or not (if relevant). Extended results for additional configurations of the methods are provided in Appendix \ref{subsec:dyn_add_results}.

\subsubsection{\emph{Lotka-Volterra} system}\label{subsubsec:lotka_volterra_experiment}
Lotka-Volterra equations model the evolution of preys $x$ and predators $y$ according to:
\begin{equation*}
    \begin{cases}
        \dot{x} &= \alpha x - \beta x y\\
        \dot{y} &= - \delta y + \gamma x y
    \end{cases}
    \quad \iff \quad
    \begin{cases}
        \dot{p} &= \alpha - \beta e^q\\
        \dot{q} &= - \delta + \gamma e^p
    \end{cases}
\end{equation*}
where $p = \ln(x)$ and $q = \ln(y)$. Due to training stability reasons, we did not manage to fit any method on the left-hand side system. Therefore, we choose to focus on the right-hand side system, where $\x = (p, q)$ is the system state. We assume $h_k^{\theta_k}(p, q) = (-\theta_k^1 e^q, 0)$ and $h_a^{\theta_a} = (\alpha, -\delta + \gamma e^p)$.

\begin{table}[h]
\caption{Results for the \emph{Lotka-Volterra} system, averaged over 5 different initializations of $h$. Reg. (resp. No Reg.) refers to using (resp. not using) the specific regularizer. Init. (resp. No Init.) refers to initializing $h_k$ as the minimizer of $d(h_k, f)$ (resp. randomly).}
\label{table:lotka_volterra_results}
\centering\footnotesize
\setlength{\tabcolsep}{3.8pt}
\begin{tabular}{lcccc}
\toprule
& \multicolumn{2}{c}{$\log$-$\hat{d}(h, f; TS)$} & \multicolumn{2}{c}{$\hat{d}_k(h_k^{\theta_k}, f_k; TS)$}\\
Method & No Reg. & Reg. & No Reg. & Reg. \\ 
\midrule 
$f_k \rightarrow h_a$ & \multicolumn{2}{c}{$-5.45 \pm 0.02$} & \multicolumn{2}{c}{-} \\ 
Joint (No Init.) & $-5.22 \pm 0.04$ & $-5.45 \pm 0.06$ & $\num{2.51e-01} \pm \num{3.90e-02}$ & $\num{1.17e-06} \pm \num{1.47e-13}$ \\ 
Alternate (Init.) & $-5.19 \pm 0.04$ & $\textbf{-5.52} \pm \textbf{0.05}$ & $\num{3.79e-01} \pm \num{4.49e-02}$ & $\num{1.35e-01} \pm \num{7.19e-03}$ \\ 
PD-based (No Init.) & \multicolumn{2}{c}{$-5.41 \pm 0.04$} & \multicolumn{2}{c}{$\textbf{\num{1.05e-07}} \pm \textbf{\num{9.51e-15}}$} \\ 
$h_a$ only & \multicolumn{2}{c}{$-5.02 \pm 0.09$} & \multicolumn{2}{c}{-} \\ 
\bottomrule
\end{tabular}
\end{table}

In terms of trajectory estimation, we can observe from \cref{table:lotka_volterra_results} that all hybrid optimization schemes perform very similarly, although alternate training performs slightly better. We can also note that all training methods yield better estimators than the purely data-driven setting. As concerns the fitting of $h_k$, results suggest that our PD-based training procedure produces a more robust and accurate estimation of $f_k$. 

We can also study the effect of the regularizers introduced by \citet{yin2021augmenting} and \citet{dona2022constrained}. For this specific problem, joint and alternate optimization combined with their regularizers seem to improve both the estimation of $f_k$ and the final predictive performance.

\subsubsection{\emph{Damped Pendulum} system}\label{subsubsec:damped_pendulum_experiment}
The evolution of the angle $\theta(t)$ of a damped pendulum system is described as follows:
\begin{equation*}
    \frac{d^2\theta}{dt^2} + \omega_0^2 \sin \theta + \xi \frac{d\theta}{dt} = 0, \quad \iff \quad \begin{cases}
        \dot{\theta} &= \frac{d\theta}{dt}\\
        \ddot{\theta} &= - \omega_0^2 \sin \theta - \xi \frac{d\theta}{dt}
    \end{cases}
\end{equation*}
where $\x = (\theta, \frac{d\theta}{dt})$ is the system state. We assume $h_k^{\theta_k}(\theta, \frac{d\theta}{dt}) = (\frac{d\theta}{dt}, - \theta_k \sin \theta)$, and $h_a^{\theta_a}$ accounts for the damping term $-\xi d\theta/dt$.

\begin{table*}[h]
\caption{Results for the \emph{Damped Pendulum} system, averaged over 5 different initializations of $h$. Reg. (resp. No Reg.) refers to using (resp. not using) the specific regularizer. Init. (resp. No Init.) refers to initializing $h_k$ as the minimizer of $d(h_k, f)$ (resp. randomly).}
\label{table:damped_pendulum_results}
\centering\footnotesize
\setlength{\tabcolsep}{3.8pt}
\begin{tabular}{lcccc}
\toprule
& \multicolumn{2}{c}{$\log$-$\hat{d}(h, f; TS)$} & \multicolumn{2}{c}{$\hat{d}_k(h_k^{\theta_k}, f_k; TS)$}\\
Method & No Reg. & Reg. & No Reg. & Reg. \\ 
\midrule 
$f_k \rightarrow h_a$ & \multicolumn{2}{c}{$\mathbf{-8.75} \pm \mathbf{0.01}$} & \multicolumn{2}{c}{-} \\ 
Joint (No Init.) & $-7.68 \pm 0.03$ & $-5.98 \pm 0.04$ & $\num{2.29e-01} \pm \num{4.47e-03}$ & $\num{5.51e-03} \pm \num{2.36e-06}$ \\ 
Alternate (Init.) & $-7.36 \pm 0.02$ & $-6.91 \pm 0.00$ & $\num{1.61e-01} \pm \num{4.09e-03}$ & $\num{5.90e-03} \pm \num{2.64e-06}$ \\ 
PD-based (No Init.) & \multicolumn{2}{c}{$-8.28 \pm 0.04$} & \multicolumn{2}{c}{$\textbf{\num{2.42e-05}} \pm \textbf{\num{4.57e-12}}$} \\ 
$h_a$ only & \multicolumn{2}{c}{$-7.28 \pm 0.01$} & \multicolumn{2}{c}{-} \\ 
\bottomrule
\end{tabular}
\end{table*}

\cref{table:damped_pendulum_results} highlights that our PD-based optimization predicts more accurate trajectories than the other hybrid training methods. Similarly, all training methods yield better estimators than the purely data-driven setting, provided we do not regularize. For this system, we see that the baseline $f_k \rightarrow h_a$ is now the optimal training approach. Concerning the estimation of $h_k$, we observe that our hybrid training scheme yields the best estimations. Contrary to the previous system, regularizing joint and alternate optimization improves the resulting $h_k$, but now degrades final predictive performance. This underlines the fact that there exists an interesting trade-off to investigate for both methods, but also that they are very sensitive to the chosen regularization hyperparameter values.

\subsubsection{\emph{Reaction-Diffusion} system}\label{subsubsec:reaction_diffusion_experiment}
This 2-dimensional system is driven by the following PDEs \citep{klaasen1984stationary}:
\begin{equation*}
    \begin{cases}
        \frac{\delta u}{\delta t} &= \alpha \Delta u + R_u(u, v; \gamma)\\
        \frac{\delta v}{\delta t} &= \beta \Delta v + R_v(u, v)\\
    \end{cases}
\end{equation*}
where $\x = (u, v)$ is the system state defined over some rectangular domain with circular boundary conditions. $\Delta$ is the Laplace operator and $\alpha, \beta$ are the diffusion coefficients for both quantities. $R_u(u, v; \gamma) = u - u^3 - \gamma - v$ and $R_v(u, v) = u - v$ are the local reaction terms. We assume $h_k^{\theta_k}(u, v) = (\theta_k^1 \Delta u, \theta_k^2 \Delta v)$, and $h_a^{\theta_a}$ accounts for the reaction terms $R_u$ and $R_v$. Contrary to the two previous experiments, due to time and computational constraints, we did not optimize hyperparameters. Instead, all training schemes were given the same set of hyperparameters that we deemed reasonable.

\begin{table}[h]
\caption{Results for the \emph{Reaction-Diffusion} system, averaged over 5 different initializations of $h$. Reg. (resp. No Reg.) refers to using (resp. not using) the specific regularizer. Init. (resp. No Init.) refers to initializing $h_k$ as the minimizer of $d(h_k, f)$ (resp. randomly).}
\label{table:reaction_diffusion_results}
\centering\footnotesize
\setlength{\tabcolsep}{2.75pt}
\begin{tabular}{lcccc}
\toprule
& \multicolumn{2}{c}{$\log$-$\hat{d}(h, f; TS)$} & \multicolumn{2}{c}{$\hat{d}_k(h_k^{\theta_k}, f_k; TS)$}\\
Method & No Reg. & Reg. & No Reg. & Reg. \\ 
\midrule 
$f_k \rightarrow h_a$ & \multicolumn{2}{c}{$-3.43 \pm 0.11$} & \multicolumn{2}{c}{-} \\ 
Joint (No Init.) & $\textbf{-3.62} \pm \textbf{0.02}$ & $-2.96 \pm 0.01$ & $\textbf{\num{1.17e-05}} \pm \textbf{\num{1.42e-10}}$ & $\num{5.45e-04} \pm \num{2.64e-07}$ \\ 
Alternate (Init.) & $-2.99 \pm 0.02$ & $-2.97 \pm 0.01$ & $\num{4.80e-05} \pm \num{2.81e-09}$ & $\num{1.89e-03} \pm \num{1.67e-06}$ \\ 
PD-based (No Init.) & \multicolumn{2}{c}{$-3.09 \pm 0.02$} & \multicolumn{2}{c}{$\num{1.64e-03} \pm \num{1.01e-07}$} \\ 
$h_a$ only & \multicolumn{2}{c}{$-2.86 \pm 0.01$} & \multicolumn{2}{c}{-} \\ 
\bottomrule
\end{tabular}
\end{table}

From \cref{table:reaction_diffusion_results}, we observe that the best performing scheme is the non-regularized joint training of $h_k$ and $h_a$, as opposed to the previous problems. As before, introducing the regularizers defined in both reference works induces a drop in final performance, still more controlled for alternate optimization. This drop is also observed in the estimation of $f_k$ (which seems optimal for the non-regularized configuration), confirming that such regularizing criteria are very sensitive to optimize.

PD-based training yields reasonable trajectory estimates, but the quality of its estimation of $h_k$ is no longer the best amongst all. Indeed, contrary to the two previous systems, it is clear that A\ref{hyp:2} is no longer satisfied as $f_a(u, v) = (u-u^3-\gamma-v, u-v)$ includes variables present in $f_k$. The latter setting represents a limitation of the PD-based optimization scheme as its estimation of the physical component is now effectively impacted, compared to the other hybrid training methods.

\section{Conclusion}
We study several methods for training hybrid additive models on static supervised regression problems, as well as on dynamical systems. On static problems, we empirically show that trends observed for neural networks also apply for the non-parametric tree-based approaches, in terms of predictive performance as well as in the estimation of the algebraic known function. We introduce claims related to the convergence of these hybrid approaches and verify their soundness on illustrative experiments. In particular, for such static additive problems where both terms are functions of disjoint and independent input features, we hypothesize that sequential and alternate training asymptotically converge to the true parametric component. 

We present a new hybrid approach leveraging partial dependence and show its competitiveness against sequential and alternate optimization schemes on both synthetic and real-world static problems. We extend the modeling framework to dynamical systems forecasting and show that PD-based training also yields robust and accurate models in such setting. We highlight its benefits in estimating the parametric prior and show that it alleviates both the risk of the ML term to dominate the known term and the need for assuming independent input features sets, although the quality of its parametric estimator may by degraded when there is a clear overlap of input features between $f_k$ and $f_a$. Another advantage of the PD-based training scheme is the reduced need for hyperparameter tuning compared to the alternate approach, which introduces multiple regularization coefficients that must be jointly optimized.

As a more general conclusion, hybrid methods are shown in our experiments to improve predictive performance with respect to ML-only models, although not always very significantly. The main benefit of the alternate and PD-based methods over the simple sequential approach is that they provide better estimators of the algebraic term, especially when filtering is relevant and can be applied. As concerns dynamical systems, such hybrid methods yield more interpretable and accurate models than purely data-driven models.

As future work, we are interested in studying theoretical conditions of optimality for our PD-based hybrid training and other approaches, in the context of dynamical systems. We also plan to try and extend current hybrid methods tackling dynamical systems forecasting so that they can be used with tree-based methods.

\backmatter

\bmhead{Acknowledgements}
This work was supported by Service Public de Wallonie Recherche under Grant No. 2010235 - ARIAC by DIGITALWALLONIA4.AI. Computational
resources have been provided by the Consortium des Equipements de Calcul Intensif (CECI), funded by the Fonds de la
Recherche Scientifique de Belgique (F.R.S.-FNRS) under Grant No. 2.5020.11 and by the Walloon Region.

\section*{Declarations}
\subsection*{Competing interests}
The authors declare they have no financial nor non-financial interests.

\bibliography{sn-bibliography}

\newpage
\begin{appendices}

\section{Optimal model under A\ref{hyp:2} and A\ref{hyp:3}}
\label{app:bayes_model_proof}

Let us recall the regression problem, where $y \in \R$ can be decomposed into the addition of two independent terms:
\begin{equation}
 y = f_k(\x_k) + f_a^r(\x_a) + \varepsilon, \quad \varepsilon \sim \mathcal{N}(0, \sigma^2), \quad \x_k \cup \x_a = \x, \x_k\cap \x_a = \emptyset, \x_k \perp\!\!\!\perp \x_a\nonumber.
\end{equation}
For clarity, let us denote by $\mathbb{E}_{\x}$ the subsequent expectations over the input space $\mathbb{E}_{\x \sim p(\x)}\left\{\cdot\right\}$. We have:
\begin{align*}
    \hat{f}_k^* &= \arg\min_{\hat{f}_k \in \hat{\mathcal{F}}_k} d(\hat{f}_k, y) = \arg\min_{\hat{f}_k \in \hat{\mathcal{F}}_k} \mathbb{E}_{\x, \varepsilon}\left\{\left(\hat{f}_k(\x_k) - f_k(\x_k) - f_a^r(\x_a) - \varepsilon\right)^2\right\}\\
    &= \arg\min_{\hat{f}_k \in \hat{\mathcal{F}}_k} \mathbb{E}_{\x_k}\left\{(\hat{f}_k(\x_k) - f_k(\x_k))^2\right\} + \mathbb{E}_{\x_a, \varepsilon}\left\{(f_a^r(\x_a) + \varepsilon)^2\right\} \\
    & - \mathbb{E}_{\x, \varepsilon}\left\{2(\hat{f}_k(\x_k) - f_k(\x_k))(f_a^r(\x_a) + \varepsilon)\right\}.
\end{align*}
The second term is independent w.r.t. $\hat{f}_k$ and thus has no impact on the minimization. Moreover, since $\x_k \perp\!\!\!\perp \x_a$, the last term writes as the product of two expectations, one of which is constant w.r.t. $\hat{f}_k$. We thus have:
\begin{equation}
\label{eq:bayes_model_final_eq}
    \hat{f}_k^* = \arg\min_{\hat{f}_k \in \hat{\mathcal{F}}_k} \mathbb{E}_{\x_k}\left\{(\hat{f}_k(\x_k) - f_k(\x_k))^2\right\} - 2 C \mathbb{E}_{\x_k}\left\{(\hat{f}_k(\x_k) - f_k(\x_k))\right\},
\end{equation}
with $C = \mathbb{E}_{\x_a, \varepsilon}\left\{(f_a^r(\x_a) + \varepsilon)\right\}  = \mathbb{E}_{\x_a}\left\{f_a^r(\x_a)\right\}$. Cancelling the derivative of \eqref{eq:bayes_model_final_eq} w.r.t. $\hat{f}_k$, we obtain the optimal model $\hat{f}_k^*(\x_k) = f_k(\x_k) + C$, for every $\x_k \in \mathcal{X}_k$.

\section{Model architectures}
\label{app:architectures}
We list below the hyperparameter values that have been used in all experiments.
\subsection{Static regression problems}
Model hyperparameters for the static regression problems are reported in \cref{tab:static_model_hyperparameters}. These values have been chosen on the basis of preliminary small-scale experiments, with the main aim of obtaining a good tradeoff between predictive performance and computation times of the full experiments. Obviously better performance can be obtained for each of the methods with a more intensive hyper-parameter tuning. We used \texttt{PyTorch} \citep{paszke2019pytorch} for MLP, \texttt{scikit-learn} \citep{scikit-learn} for RF and \texttt{xgboost} \citep{chen2016xgboost} for GB. Unspecified parameters keep their default values. Learning rates for training $h_k$ and MLP are set to $0.005$. $\operatorname{H}$ is the number of hidden layers in MLP and $\operatorname{W}$ the number of neurons per hidden layer.
$\operatorname{T}$ is the number of trees in GB and RF, $\operatorname{d}$ the maximum tree depth and $\operatorname{mss}$ the minimum number of samples required to split an internal tree node.
\begin{table}[h]
    \centering
    \caption{Model hyperparameters, for each experiment.}
    \label{tab:static_model_hyperparameters}
    \setlength{\tabcolsep}{8pt}
    \begin{tabular}{lccc}
    \toprule
    Problem & MLP & GB & RF \\
    \hline
    \emph{(Correlated) Friedman} problems & $\operatorname{H}=2, \operatorname{W}=15$ & $\operatorname{T}=700, \operatorname{d}=2$ & $\operatorname{T}=500, \operatorname{mss}=5$ \\
    \hline
    \emph{Linear} \& \emph{Overlapping} problems & $\operatorname{H}=2, \operatorname{W}=10$ & $\operatorname{T}=400, \operatorname{d}=2$ & $\operatorname{T}=500, \operatorname{mss}=5$ \\
    \hline
    Real-world problems & $\operatorname{H}=2, \operatorname{W}=30$ & $\operatorname{T}=300, \operatorname{d}=2$ & $\operatorname{T}=200, \operatorname{mss}=5$ \\
    \bottomrule
    \end{tabular}
\end{table}

\subsection{Dynamical systems forecasting}
Model hyperparameters for the dynamical forecasting problems are reported in Tables \ref{tab:lotka_volterra_model_hyperparameters}, \ref{tab:pendulum_model_hyperparameters} and \ref{tab:rd_model_hyperparameters}. We used \texttt{PyTorch} for MLP and CNN. Tested hyperparameters are reported in \cref{tab:hyperparameter_values}. Given that we only optimized hyperparameters for the \emph{Lotka-Volterra} and \emph{Damped Pendulum} systems, architectural hyperparameters solely refer to those of MLPs. According to the algorithm described in \citep{yin2021augmenting} and \citep{dona2022constrained}, $\lambda_h$ is optimized for joint optimization whereas $\lambda_h$, $\lambda_{h_k}$ and $\lambda_{h_a}$ are optimized for alternate training.

\begin{table}
    \caption{Hyperparameter values tested for the \emph{Lotka-Volterra} and \emph{Damped Pendulum} systems.}
    \label{tab:hyperparameter_values}
    \centering
    \begin{tabular}{lccc}
    \toprule
    Hyperparameter &  &  &  \\
    \hline
    Learning Rate (LR) & $\num{5e-4}$ & $\num{1e-4}$ & $\num{1e-5}$ \\
    Num. hidden layers & 1 & 2 & - \\
    Hidden size & 16 & 32 & 64 \\
    $\lambda_h$ - Joint & $1$ & $10$ & $20$ \\
    $\lambda_h$ - Alternate & $\num{1e-1}$ & $\num{1e-2}$ & $\num{1e-3}$ \\
    $\lambda_{h_k}$ - Alternate & $\num{1e-1}$ & $\num{1e-2}$ & $\num{1e-3}$ \\
    $\lambda_{h_a}$ - Alternate & $\num{1e-1}$ & $\num{1e-2}$ & $\num{1e-3}$ \\
    \bottomrule
    \end{tabular}
\end{table}

For each problem, for a given optimization scheme and for each of the five different initializations of $h$, we tested every combination of hyperparameters and evaluated it on the corresponding validation set. As optimal hyperparameters may vary from one seed to the other, we checked all five optimal configurations of the given training method and reported the most frequent one. When no clear majority could be established, we reported several optimal values for the same hyperparameter. Note that the optimal hyperparameters across the five seeds were most of the time identical.

\begin{table}[h]
    \caption{Optimal model hyperparameters (MLP), for the \emph{Lotka-Volterra} system.}
    \label{tab:lotka_volterra_model_hyperparameters}
    \centering\footnotesize
    \setlength{\tabcolsep}{6pt}
    \begin{tabular}{lcccccc}
    \toprule
    Method & LR & Num. layers & Hidden size & $\lambda_h$ & $\lambda_{h_k}$ & $\lambda_{h_a}$ \\
    \hline
    $f_k \rightarrow h_a$ & $\num{5e-4}$ & 2 & 64 & - & - & - \\
    Joint - No Init. \& No Reg. & $\num{5e-4}$ & 2 & 64 & - & - & - \\
    Joint - No Init. \& Reg. & $\num{1e-5}$ & 2 & 64 & 10/20 & - & -\\
    Joint - Init. \& No Reg. & $\num{5e-4}$ & 2 & 64 & - & - & - \\
    Joint - Init. \& Reg. & $\num{1e-5}$ & 2 & 64 & 20 & - & - \\
    Alternate - No Init. \& No Reg. & $\num{5e-4}$ & 2 & 64 & - & - & - \\
    Alternate - No Init. \& Reg. & $\num{5e-4}$ & 2 & 64 & $0.1$ & $0.001$ & $0.001$ \\
    Alternate - Init. \& No Reg. & $\num{5e-4}$ & 2 & 64 & - & - & - \\
    Alternate - Init. \& Reg. & $\num{5e-4}$ & 2 & 64 & $0.1$ & $0.01$ & $0.001$ \\
    PD-based - No Init. & $\num{5e-4}$ & 2 & 64 & - & - & - \\
    PD-based - Init. & $\num{5e-4}$ & 2 & 64 & - & - & - \\
    $h_a$ only & $\num{5e-4}$ & 2 & 64 & - & - & -\\
    \bottomrule
    \end{tabular}
\end{table}

\begin{table}[h]
    \centering
    \caption{Optimal model hyperparameters (MLP), for the \emph{Damped Pendulum} system.}
    \label{tab:pendulum_model_hyperparameters}
    \centering\footnotesize
    \setlength{\tabcolsep}{6pt}
    \begin{tabular}{lcccccc}
    \toprule
    Method & LR & Num. layers & Hidden size & $\lambda_h$ & $\lambda_{h_k}$ & $\lambda_{h_a}$ \\
    \hline
    $f_k \rightarrow h_a$ & $\num{5e-4}$ & 2 & 64 & - & - & - \\
    Joint - No Init. \& No Reg. & $\num{5e-4}$ & 2 & 64 & - & - & - \\
    Joint - No Init. \& Reg. & $\num{1e-4}$ & 1/2 & 32/16 & 20 & - & - \\
    Joint - Init. \& No Reg. & $\num{5e-4}$ & 2 & 64 & - & - & - \\
    Joint - Init. \& Reg. & $\num{1e-4}$ & 1/2 & 16 & 20 & - & - \\
    Alternate - No Init. \& No Reg. & $\num{5e-4}$ & 2 & 64 & - & - & - \\
    Alternate - No Init. \& Reg. & $\num{5e-4}$ & 2 & 64 & $0.1$ & $0.001$ & $0.001$ \\
    Alternate - Init. \& No Reg. & $\num{5e-4}$ & 2 & 64 & - & - & - \\
    Alternate - Init. \& Reg. & $\num{5e-4}$ & 2 & 64 & $0.1$ & $0.001$ & $0.001$ \\
    PD-based - No Init. & $\num{5e-4}$ & 2 & 64 & - & - & - \\
    PD-based - Init. & $\num{5e-4}$ & 2 & 64 & - & - & - \\
    $h_a$ only & $\num{5e-4}$ & 2 & 64 & - & - & - \\
    \bottomrule
    \end{tabular}
\end{table}

\begin{table}[h]
    \centering
    \caption{Model hyperparameters (CNN), for the \emph{Reaction-Diffusion} system.}
    \label{tab:rd_model_hyperparameters}
    \centering\footnotesize
    \setlength{\tabcolsep}{6pt}
    \begin{tabular}{lcccccc}
    \toprule
    Method & LR & Num. layers & Hidden channels  & $\lambda_h$ & $\lambda_{h_k}$ & $\lambda_{h_a}$ \\
    \hline
    $f_k \rightarrow h_a$ & $\num{1e-4}$ & 3 & 8 & - & - & - \\
    Joint - No Init. \& No Reg. & $\num{1e-4}$ & 3 & 8 & - & - & - \\
    Joint - No Init. \& Reg. & $\num{1e-4}$ & 3 & 8 & 20 & - & - \\
    Joint - Init. \& No Reg. & $\num{1e-4}$ & 3 & 8 & - & - & -\\
    Joint - Init. \& Reg. & $\num{1e-4}$ & 3 & 8 & 20 & - & - \\
    Alternate - No Init. \& No Reg. & $\num{1e-4}$ & 3 & 8 & - & - & - \\
    Alternate - No Init. \& Reg. & $\num{1e-4}$ & 3 & 8 & $0.1$ & $0.01$ & $0.001$ \\
    Alternate - Init. \& No Reg. & $\num{1e-4}$ & 3 & 8 & - & - & - \\
    Alternate - Init. \& Reg. & $\num{1e-4}$ & 3 & 8 & $0.1$ & $0.01$ & $0.001$ \\
    PD-based - No Init. & $\num{1e-4}$ & 3 & 8 & - & - & - \\
    PD-based - Init. & $\num{1e-4}$ & 3 & 8 & - & - & - \\
    $h_a$ only & $\num{1e-4}$ & 3 & 8 & - & - & - \\
    \bottomrule
    \end{tabular}
\end{table}

\section{Impact of training sample size - Additional results}
\label{app:training_size_additional}
\begin{figure}[h]
    \centering
    \includegraphics[width=\linewidth]{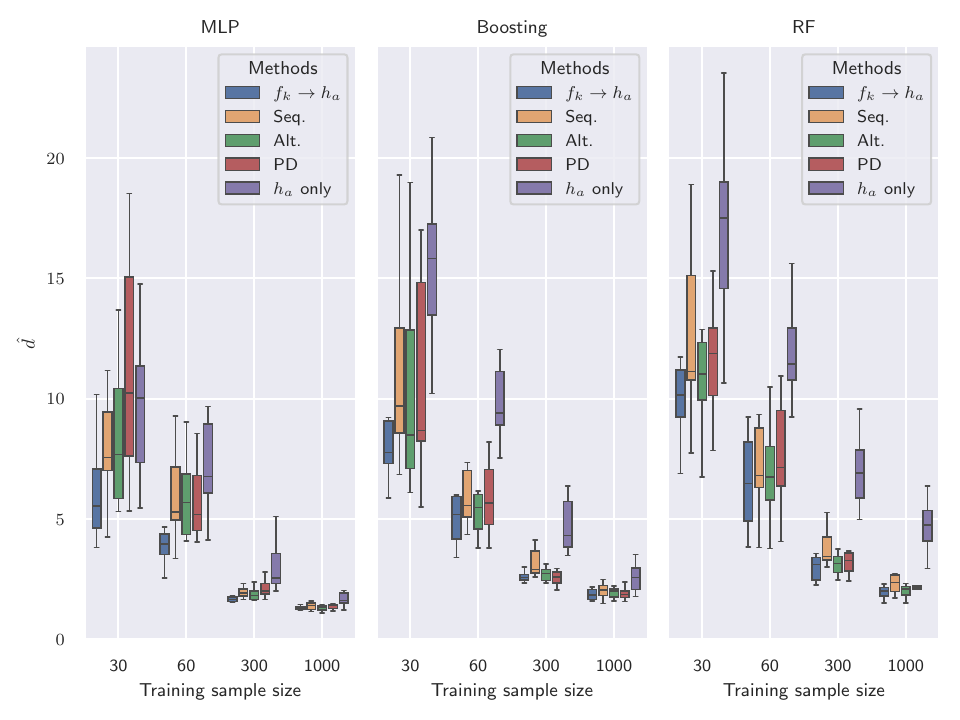}
    \vspace{-4ex}
    \caption{Evolution of $\hat{d}$  w.r.t. the training sample size, on the \emph{Correlated Friedman} problem. The boxplot summarizes the results over 10 test sets.}
    \label{fig:impact_ts_mse_all}
\end{figure}

\begin{figure}[h]
    \centering
    \includegraphics[width=\linewidth]{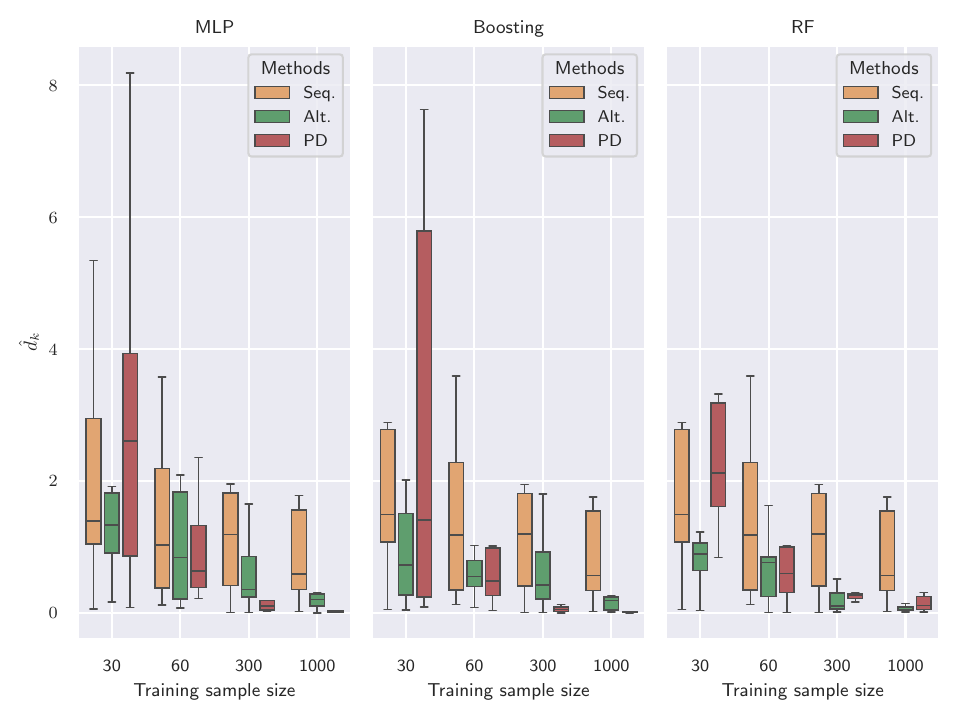}
    \vspace{-4ex}
    \caption{Evolution of $\hat{d}_k$ w.r.t. the training sample size, on the \emph{Correlated Friedman} problem. The boxplot summarizes the results over 10 test sets.}
    \label{fig:impact_ts_fp_mse_all}
\end{figure}

Similarly to MLP, we observe in \cref{fig:impact_ts_mse_all} and \cref{fig:impact_ts_fp_mse_all} that hybrid methods with boosting outperform data-driven methods. Once again, PD-based training seems to struggle with very low sample sizes, although the gap with other hybrid schemes is more moderate and the performance of boosting alone is worse than that of MLP alone. Whatever the training sample size, the performance of hybrid methods with boosting is on par (and sometimes better) with that of MLP, both for generalization error and estimates of $h_k$.

The same observations can be made for RF, even though its performance is slightly worse than its MLP or boosting counterparts, whatever the training sample size. Still, it is interesting to note that hybrid methods using RF yield the best estimates of $h_k$ for the smallest sample size.

\section{Dynamical experiments details}
\label{app:dynamical_details}
\subsection{\emph{Lotka-Volterra} system}
The system is driven by the following dynamics:
\begin{equation*}
    \begin{cases}
        \dot{x} &= \alpha x - \beta x y\\
        \dot{y} &= - \delta y + \gamma x y
    \end{cases}
\end{equation*}
where prey and predator quantities $x$ and $y$ can be transformed as $p = \ln(x), q = \ln(y)$ such that
\begin{equation*}
    \begin{cases}
        \dot{p} &= \alpha - \beta e^q\\
        \dot{q} &= - \delta + \gamma e^p
    \end{cases}
\end{equation*}
where $\x = (p, q)$ is the system state. 

We simulate a dataset of 200 trajectories spanning 401 timesteps (including the initial state $\x_0$), over a time interval of $[0, 20]$s. We first generate initial prey-predator proportions $(x, y) \sim \mathcal{U}[0, 1]^2$ and compute the resulting initial states $\x_0 = (\ln(x), \ln(y))$. We set $\alpha = \beta = \gamma = \delta = 1$ and generate trajectories by integrating for 400 steps ($\delta t = 0.05$) using a 4th order Runge-Kutta scheme, with an integration timestep $\delta t_{sim} = 0.001$. We split the dataset into train-validation-test subsets (50/25/25\%). The train set is eventually obtained by extracting windows of 41 timesteps (with a stride of 2) from the original train split, yielding $18\,100$ smaller sub-trajectories.

For the data-driven, joint and alternate schemes, models are trained for 500 epochs using the Euler integrator with $\delta t = 0.05$. For the PD-based optimization scheme, each training consists of 50 epochs repeated 10 times, with the same integration settings. The final training of $h_a$ consists of 150 epochs. All methods are optimized with Adam \citep{kingma2014adam}.

\subsection{\emph{Damped Pendulum} system}
The damped pendulum dynamics are described by:
\begin{equation*}
    \begin{cases}
        \dot{\theta} &= \frac{d\theta}{dt}\\
        \ddot{\theta} &= - \omega_0^2 \sin \theta - \xi \frac{d\theta}{dt}
    \end{cases}
\end{equation*}
where $\x = (\theta, \frac{d\theta}{dt})$ is the system state.

We simulate a dataset of 200 trajectories spanning 201 timesteps (including the initial state $\x_0$), over a time interval of $[0, 10]$s. We first generate the initial states $\x_0 = (\theta_0, \frac{d\theta_0}{dt}) \sim \mathcal{U}[\frac{-\pi}{2}, \frac{pi}{2}] \times \mathcal{U}[0, 0.1]$. We set $(\omega, \xi) \sim \mathcal{U}[0.785, 3.14] \times \mathcal{U}[0, 0.8]$ and generate trajectories by integrating for 200 steps ($\delta t = 0.05$) using a 4th order Runge-Kutta scheme, with an integration timestep $\delta t_{sim} = 0.001$. We split the dataset into train-validation-test subsets (50/25/25\%). The train set is eventually obtained by extracting windows of 41 timesteps (with a stride of 2) from the original train split, yielding $8\,100$ smaller sub-trajectories.

For the data-driven, joint and alternate schemes, models are trained for 500 epochs using the Euler integrator with $\delta t = 0.05$. For the PD-based optimization scheme, each training consists of 50 epochs repeated 10 times, with the same integration settings. The final training of $h_a$ consists of 150 epochs. All methods are optimized with Adam \citep{kingma2014adam}.

\subsection{\emph{Reaction-Diffusion} system}
The reaction-diffusion system is a 2-dimensional system driven by the following PDEs \citep{klaasen1984stationary}:
\begin{equation*}
    \begin{cases}
        \frac{\delta u}{\delta t} &= \alpha \Delta u + R_u(u, v; \gamma)\\
        \frac{\delta v}{\delta t} &= \beta \Delta v + R_v(u, v)\\
    \end{cases}
\end{equation*}
where $\x = (u, v)$ is the system state. $R_u(u, v; \gamma) = u - u^3 - \gamma - v$ and $R_v(u, v) = u - v$ are the local reaction terms. We choose a rectangular domain spanning $[-1, 1]^2$ with circular boundary conditions, that we discretize by a $32 \times 32$ uniform grid $G(t)$, where $\Delta$ is the $3 \times 3$ Laplace operator. For this system, $(u,v)$ values are associated to each of the $32 \times 32$ grid cells $G_{ij}$ that are governed by the same PDEs. Therefore, we decided to model $f_a = (R_u(u, v; \gamma), R_v(u, v))$ through a CNN with input $G(t)$.

For the generation of the dataset, we take inspiration from the work of \citet{yin2021augmenting}. We simulate a dataset of $1\,920$ trajectories spanning 246 timesteps (including the initial state $\x_0$), over a time interval of $[0, 2.45]$s. We first generate the initial states $\x_{0} = (u_{0}, v_{0}) \sim \mathcal{U}[0, 1]^2$. We set $\alpha = 1\mathrm{e}{-3}, \beta = 5\mathrm{e}{-3}, \gamma = 5\mathrm{e}{-3}$ and generate trajectories by integrating for 245 steps ($\delta t = 0.01$) using a 4th order Runge-Kutta scheme, with an integration timestep $\delta t_{sim} = 0.001$. We split the dataset into train-validation-test subsets (50/25/25\%). The train set is eventually obtained by extracting windows of 51 timesteps (with a stride of 20) from the original train split, yielding $9\,600$ smaller sub-trajectories.

For the data-driven, joint and alternate schemes, models are trained for $1\,000$ epochs using the Euler integrator with $\delta t = 0.01$. For the PD-based optimization scheme, each training consists of 100 epochs repeated 10 times, with the same integration settings. The final training of $h_a$ consists of $1\,000$ epochs. All methods are optimized with Adam \citep{kingma2014adam}.

\subsection{Dynamical systems forecasting - Additional results}\label{subsec:dyn_add_results}
In this section, we report results for variants of the optimization methods presented in \cref{subsec:dyn_methods}. In particular, we test the initialization introduced by \citet{dona2022constrained} with both joint and PD-based training. For PD-based training, this means replacing the first two steps of Algorithm \ref{alg:pdp_opt_general_dynamic} with a fit of $(\hat{\theta}_k,\hat{\gamma})$ to minimize $d(h_k^{{\theta}_k}+\gamma, f)$. We also compare the impact of regularizing $h_a$ for both joint and alternate optimizations.

\subsubsection{\emph{Lotka-Volterra} system}

\begin{table}[h]
    \centering
    \caption{Results for other \emph{Lotka-Volterra} system, averaged over 5 different initializations of $h$. Init. refers to $(\hat{\theta}_k, \hat{\gamma}) = \arg\min_{\theta_k, \gamma}d(h_k^{\theta_k} + \gamma, f)$ while Reg. (resp. No Reg.) refers to using (resp. not using) the specific regularizer. Methods not included in the table in the main paper are in italics.}
    \label{tab:full_lotka_volterra_results}
    \setlength{\tabcolsep}{8pt}
    \begin{tabular}{lcc}
    \toprule
    Method & $\log$-$\hat{d}(h, f; TS)$ & $\hat{d}_k(h_k^{\theta_k}, f_k; TS)$ \\
    \hline
    $f_k \rightarrow h_a$ & $-5.45 \pm 0.02$ & -\\
    Joint - No Init. \& No Reg. & $-5.22 \pm 0.04$ & $\num{2.51e-01} \pm \num{3.90e-02}$ \\
    Joint - No Init. \& Reg. & $-5.45 \pm 0.06$ & $\num{1.17e-06} \pm \num{1.47e-13}$ \\
    \emph{Joint - Init. \& No Reg.} & $\textbf{-5.53} \pm \textbf{0.01}$ & $\num{8.62e-02} \pm \num{2.63e-03}$ \\
    \emph{Joint - Init. \& Reg.} & $-5.42 \pm 0.04$ & $\num{1.66e-06} \pm \num{7.63e-13}$ \\
    \emph{Alternate - No Init. \& No Reg.} & $-5.23 \pm 0.09$ & $\num{5.58e-01} \pm \num{1.64e-01}$ \\
    \emph{Alternate - No Init. \& Reg.} & $-5.32 \pm 0.01$ & $\num{1.66e-01} \pm \num{5.00e-04}$ \\
    Alternate - Init. \& No Reg. & $-5.19 \pm 0.04$ & $\num{3.79e-01} \pm \num{4.49e-02}$ \\
    Alternate - Init. \& Reg. & $\textbf{-5.52} \pm \textbf{0.05}$ & $\num{1.35e-01} \pm \num{7.19e-03}$ \\
    PD-based - No Init. & $-5.41 \pm 0.04$ & $\num{1.05e-07} \pm \num{9.51e-15}$ \\
    \emph{PD-based - Init.} & $\textbf{-5.51} \pm \textbf{0.02}$ & $\textbf{\num{6.27e-08}} \pm \textbf{\num{8.72e-15}}$ \\
    $h_a$ only & $-5.02 \pm 0.09$ & - \\
    \bottomrule
    \end{tabular}
\end{table}

Compared to \cref{subsubsec:lotka_volterra_experiment}, \cref{tab:full_lotka_volterra_results} informs us that, for this problem, PD-based training can yield models equally good to alternate optimization, provided we initialize $h_k$ as specifically defined by \citet{dona2022constrained}. We can also note that the resulting estimation of $h_k$ is even better. Moreover, joint training with no regularization reaches similar predictive performance, at the cost of a worse estimation of $h_k$ (although better than that of alternate training). Given that the parametric component we fit is $h_k^{\theta_k} + \gamma$ (with $\gamma \in \mathbb{R}$), initializing $h_k$ as $(\hat{\theta}_k, \hat{\gamma}) = \arg\min_{\theta_k, \gamma}d(h_k^{\theta_k} + \gamma, f)$ actually makes $\gamma$ converge to $\alpha$, which explains why training methods seem to benefit from this initialization in this particular system.

We also investigated the impact of removing this specific $h_k$ initialization from the alternate scheme. For this problem, it appeared clearly that such initialization enhanced the final model $h$.

\subsubsection{\emph{Damped Pendulum} system}
\begin{table}[h]
    \centering
    \caption{Results for the \emph{Damped Pendulum} system, averaged over 5 different initializations of $h$. Init. refers to $(\hat{\theta}_k, \hat{\gamma}) = \arg\min_{\theta_k, \gamma}d(h_k^{\theta_k} + \gamma, f)$ while Reg. (resp. No Reg.) refers to using (resp. not using) the specific regularizer. Methods not included in the table in the main paper are in italics.}
    \label{tab:full_damped_pendulum_results}
    \setlength{\tabcolsep}{8pt}
    \begin{tabular}{lcc}
    \toprule
    Method & $\log$-$\hat{d}(h, f; TS)$ & $\hat{d}_k(h_k^{\theta_k}, f_k; TS)$ \\
    \hline
    $f_k \rightarrow h_a$ & $\textbf{-8.75} \pm \textbf{0.01}$ & -\\
    Joint - No Init. \& No Reg. & $-7.68 \pm 0.03$ & $\num{2.29e-01} \pm \num{4.47e-03}$ \\
    Joint - No Init. \& Reg. & $-5.98 \pm 0.04$ & $\num{5.51e-03} \pm \num{2.36e-06}$ \\
    \emph{Joint - Init. \& No Reg.} & $-7.74 \pm 0.02$ & $\num{1.57e-01} \pm \num{3.75e-03}$ \\
    \emph{Joint - Init. \& Reg.} & $-5.83 \pm 0.03$ & $\num{2.83e-02} \pm \num{2.26e-03}$ \\
    \emph{Alternate - No Init. \& No Reg.} & $-7.23 \pm 0.04$ & $\num{2.34e-01} \pm \num{4.47e-03}$ \\
    \emph{Alternate - No Init.\& Reg.} & $-6.91 \pm 0.00$ & $\num{6.84e-03} \pm \num{4.56e-06}$ \\
    Alternate - Init. \& No Reg. & $-7.36 \pm 0.02$ & $\num{1.61e-01} \pm \num{4.09e-03}$ \\
    Alternate - Init. \& Reg. & $-6.90 \pm 0.00$ & $\num{5.90e-03} \pm \num{2.64e-06}$ \\
    PD-based - No Init. & $-8.28 \pm 0.04$ & $\textbf{\num{2.42e-05}} \pm \textbf{\num{4.57e-12}}$ \\
    \emph{PD-based - Init.} & $-8.29 \pm 0.02$ & $\textbf{\num{2.36e-05}} \pm \textbf{\num{5.56e-12}}$ \\
    $h_a$ only & $-7.28 \pm 0.01$ & - \\
    \bottomrule
    \end{tabular}
\end{table}

Similarly to \cref{subsubsec:damped_pendulum_experiment}, we can see from \cref{tab:full_damped_pendulum_results} that PD-based training reaches the highest performance both for trajectory estimation as well as for the estimation of $h_k$. For this system, the particular $h_k$ initialization does not seem to impact our training method, on any level.

There is now more nuance to the impact of such initialization on joint optimization. Indeed, while it seems to help the non-regularized configuration, the effect is the opposite when we add the presence of their regularizer, indicating that hyperparameters could have been optimized even more or simply that their regularization criterion is not suited with this specific initialization. However, we can observe that the effect is positive for models optimized in an alternate fashion.

\subsubsection{\emph{Reaction-Diffusion} system}
\begin{table}[h]
    \centering
    \caption{Results for the \emph{Reaction-Diffusion} system, averaged over 5 different initializations of $h$. Init. refers to $(\hat{\theta}_k, \hat{\gamma}) = \arg\min_{\theta_k, \gamma}d(h_k^{\theta_k} + \gamma, f)$ while Reg. (resp. No Reg.) refers to using (resp. not using) the specific regularizer. Methods not included in the table in the main paper are in italics.}
    \label{tab:full_reaction_diffusion_results}
    \setlength{\tabcolsep}{8pt}
    \begin{tabular}{lcc}
    \toprule
    Method & $\log$-$\hat{d}(h, f; TS)$ & $\hat{d}_k(h_k^{\theta_k}, f_k; TS)$ \\
    \hline
    $f_k \rightarrow h_a$ & $-3.43 \pm 0.11$ & -\\
    Joint - No Init. \& No Reg. & $\textbf{-3.62} \pm \textbf{0.02}$ & $\textbf{\num{1.17e-05}} \pm \textbf{\num{1.42e-10}}$ \\
    Joint - No Init. \& Reg. & $-2.96 \pm 0.01$ & $\num{5.45e-04} \pm \num{2.64e-07}$ \\
    \emph{Joint - Init. \& No Reg.} & $-3.14 \pm 0.11$ & $\num{4.43e-05} \pm \num{2.61e-09}$ \\
    \emph{Joint - Init. \& Reg.} & $-2.65 \pm 0.04$ & $\num{1.14e-03} \pm \num{2.50e-07}$ \\
    \emph{Alternate - No Init. \& No Reg.} & $-3.26 \pm 0.02$ & $\num{1.94e-04} \pm \num{6.95e-09}$ \\
    \emph{Alternate - No Init. \& Reg.} & $-2.98 \pm 0.02$ & $\num{1.71e-04} \pm \num{6.18e-09}$ \\
    Alternate - Init. \& No Reg. & $-2.99 \pm 0.02$ & $\num{4.80e-05} \pm \num{2.81e-09}$ \\
    Alternate - Init. \& Reg. & $-2.97 \pm 0.01$ & $\num{1.89e-03} \pm \num{1.67e-06}$ \\
    PD-based - No Init. & $-3.09 \pm 0.02$ & $\num{1.64e-03} \pm \num{1.01e-07}$ \\
    \emph{PD-based - Init.} & $-3.12 \pm 0.02$ & $\num{2.66e-03} \pm \num{2.51e-06}$ \\
    $h_a$ only & $-2.86 \pm 0.01$ & - \\
    \bottomrule
    \end{tabular}
\end{table}

For this 2-D problem, \cref{tab:full_reaction_diffusion_results} suggests that non-regularized joint optimization with uniform $h_k$ initialization performs best, as could be seen from \cref{table:reaction_diffusion_results}. As was said, PD-based training seems to struggle now that there is a clear overlap between $f_k$ and $f_a$. Furthermore, the particular $h_k$ initialization does not seem to impact our training method.

As concerns joint optimization, the initialization proposed by \citet{dona2022constrained} clearly degrades both performance criteria. Nevertheless, we can observe that once again the effect is positive for regularized alternate training, indicating that, as could be observed from previous systems, there is a very sensitive balance to find between the initialization of the physical component and the regularizers, along with their optimal coefficient values.

\section{Training methods pseudo-codes}\label{sec:pseudocodes}
\begin{algorithm}
    \caption{Sequential Optimization: Static problems}
    \label{alg:sequential_opt_general_static}
    \begin{algorithmic}
        \State {\bfseries Input:} LS = $\{(\x_i, y_i)\}_{i=1}^N$
        \State 
        \State $(\hat{\theta}_k, \hat{\gamma}) \leftarrow \operatorname{fit}^{h_k+\gamma}(LS)$
        \State $\hat{\theta}_a \leftarrow \operatorname{fit}^{h_a}(\{(\x_i, y_i - h_k^{\hat{\theta}_k}(\x_i)-\hat{\gamma})\}_{i=1}^N)$
        \State \Return $h_k^{\hat{\theta}_k}+\hat{\gamma}+h_a^{\hat{\theta}_a}$
    \end{algorithmic}
\end{algorithm}

\begin{algorithm}
    \caption{Joint Optimization: Non-Regularized Setting - Dynamical problems}
    \label{alg:joint_opt_general}
    \begin{algorithmic}
        \State {\bfseries Input:} LS = $\{(\x_0^{(i)}, \dots, \x_T^{(i)})\}_{i=1}^N$, $\eta>0$
        \State
        \State $\hat{\theta}_k^{(0)}, \hat{\gamma}^{(0)}, \hat{\theta}_a^{(0)} \leftarrow$ Random Initialization
        \For{$e=1$ {\bfseries to} $N_{e}$}
        \State $(\hat{\theta}_k^{(e)}, \hat{\gamma}^{(e)}, \hat{\theta}_a^{(e)}) \leftarrow (\hat{\theta}_k^{(e-1)}, \hat{\gamma}^{(e-1)}, \hat{\theta}_a^{(e-1)}) - \eta \nabla_{(\hat{\theta}_k, \hat{\gamma}, \hat{\theta}_a)}[\hat{d}(h_k^{\hat{\theta}_k^{(e-1)}} + \hat{\gamma}^{(e-1)} + h_a^{\hat{\theta}_a^{(e-1)}}, f;LS)]$
        \EndFor
        \State \Return $h_k^{\hat{\theta}_k^{(N_e)}}+\hat{\gamma}^{(N_e)}+h_a^{\hat{\theta}_a^{(N_e)}}$
    \end{algorithmic}
\end{algorithm}

\begin{algorithm}
    \caption{Alternate Optimization: Static problems}
    \label{alg:alternating_opt_general_static}
    \begin{algorithmic}
        \State {\bfseries Input:} LS = $\{(\x_i, y_i)\}_{i=1}^N$, $\eta>0$
        \State
        \State $\hat{\theta}_a^{(0)} \leftarrow$ Random Initialization
        \State  $(\hat{\theta}_k^{(0)}, \hat{\gamma}^{(0)}) \leftarrow \operatorname{fit}^{h_k+\gamma}(LS)$
        \For{$e=1$ {\bfseries to} $N_{e}$}
        \State $\hat{\theta}_a^{(e)} \leftarrow \operatorname{fit}^{h_a}(\{(\x_i, y_i - h_k^{\hat{\theta}_k^{(e-1)}}(\x_i)-\hat{\gamma}^{(e-1)})\}_{i=1}^N)$
        \State $(\hat{\theta}_k^{(e)}, \hat{\gamma}^{(e)}) \leftarrow (\hat{\theta}_k^{(e-1)}, \hat{\gamma}^{(e-1)}) - \eta \nabla_{(\hat{\theta}_k, \hat{\gamma})}[\hat{d}(h_k^{\hat{\theta}_k^{(e-1)}} + \hat{\gamma}^{(e-1)} + h_a^{\hat{\theta}_a^{(e)}}, y;LS)]$
        \EndFor
        \State $\hat{\theta}_a^{(N_e+1)} \leftarrow \operatorname{fit}^{h_a}(\{(\x_i, y_i - h_k^{\hat{\theta}_k^{(N_e)}}(\x_i)-\hat{\gamma}^{(N_e)})\}_{i=1}^N)$
        \State \Return $h_k^{\hat{\theta}_k^{(N_e)}}+\hat{\gamma}^{(N_e)}+h_a^{\hat{\theta}_a^{(N_e+1)}}$
    \end{algorithmic}
\end{algorithm}

\begin{algorithm}
    \caption{Alternate Optimization: Non-Regularized Setting - Dynamical problems}
    \label{alg:alternating_opt_general_dynamic}
    \begin{algorithmic}
        \State {\bfseries Input:} LS = $\{(\x_0^{(i)}, \dots, \x_T^{(i)})\}_{i=1}^N$, $\eta>0$
        \State
        \State $\hat{\theta}_a^{(0)} \leftarrow$ Random Initialization
        \State  $(\hat{\theta}_k^{(0)}, \hat{\gamma}^{(0)}) \leftarrow \operatorname{fit}_{\operatorname{NODE}}^{h_k+\gamma}(LS)$
        \For{$e=1$ {\bfseries to} $N_{e}$}
        \State $\hat{\theta}_a^{(e)} \leftarrow \hat{\theta}_a^{(e-1)} - \eta \nabla_{\hat{\theta}_a}[\hat{d}(h_a^{\hat{\theta}_a^{(e-1)}} + h_k^{\hat{\theta}_k^{(e-1)}} + \hat{\gamma}^{(e-1)}, f;LS)]$
        \State $(\hat{\theta}_k^{(e)}, \hat{\gamma}^{(e)}) \leftarrow (\hat{\theta}_k^{(e-1)}, \hat{\gamma}^{(e-1)}) - \eta \nabla_{(\hat{\theta}_k, \hat{\gamma})}[\hat{d}(h_k^{\hat{\theta}_k^{(e-1)}} + \hat{\gamma}^{(e-1)} + h_a^{\hat{\theta}_a^{(e)}}, f;LS)]$
        \EndFor
        \State $\hat{\theta}_a^{(N_e+1)} \leftarrow \hat{\theta}_a^{(N_e)} - \eta \nabla_{\hat{\theta}_a}[\hat{d}(h_a^{\hat{\theta}_a^{(N_e)}} + h_k^{\hat{\theta}_k^{(N_e)}} + \hat{\gamma}^{(N_e)}, f;LS)]$
        \State \Return $h_k^{\hat{\theta}_k^{(N_e)}}+\hat{\gamma}^{(N_e)}+h_a^{\hat{\theta}_a^{(N_e+1)}}$
    \end{algorithmic}
\end{algorithm}

\end{appendices}

\end{document}